\PassOptionsToPackage{table}{xcolor}
\documentclass[10pt,twocolumn,letterpaper]{article}

\usepackage{iccv}              

\definecolor{iccvblue}{rgb}{0.21,0.49,0.74}
\usepackage[pagebackref,breaklinks,colorlinks,allcolors=iccvblue]{hyperref}
\usepackage{times}
\usepackage{url}
\usepackage{booktabs}
\usepackage{multirow, multicol}
\usepackage{makecell}
\usepackage{subcaption}
\usepackage{graphicx}
\usepackage{bbding}
\usepackage{pifont}
\usepackage{alltt}
\usepackage{algorithm}
\usepackage{algorithmic}
\usepackage{wrapfig, lipsum}
\usepackage{caption}
\usepackage[accsupp]{axessibility}

\definecolor{codegreen}{rgb}{0.0, 0.6, 0.0}
\definecolor{codered}{rgb}{0.6, 0.0, 0.0}
\definecolor{codeblue}{rgb}{0.2, 0.2, 1.0}
\definecolor{lightcyan}{rgb}{0.83, 0.94, 0.98}
\definecolor{sgray}{HTML}{EDEDED}

\def\paperID{6770} 
\def\confName{ICCV}
\def\confYear{2025}

\title{VRM: Knowledge Distillation via Virtual Relation Matching}

\author{Weijia Zhang$^{1}$ ~~~~~~~~~~ Fei Xie$^{1}$ ~~~~~~~~~~ Weidong Cai$^{2}$ ~~~~~~~~~~ Chao Ma$^{1}$\thanks{Corresponding author}\\
$^{1}$Shanghai Jiao Tong University ~~~~ $^{2}$The University of Sydney \\
{\tt\small \{weijia.zhang, feixie, chaoma\}@sjtu.edu.cn ~ tom.cai@sydney.edu.au}
}

\begin{document}
\maketitle
\begin{abstract}
Knowledge distillation (KD) aims to transfer the knowledge of a more capable yet cumbersome teacher model to a lightweight student model. In recent years, relation-based KD methods have fallen behind, as their instance-matching counterparts dominate in performance. In this paper, we revive relational KD by identifying and tackling several key issues in relation-based methods, including their susceptibility to overfitting and spurious responses. Specifically, we transfer novelly constructed affinity graphs that compactly encapsulate a wealth of beneficial inter-sample, inter-class, and inter-view correlations by exploiting virtual views and relations as a new kind of knowledge. As a result, the student has access to richer guidance signals and stronger regularisation throughout the distillation process. 
To further mitigate the adverse impact of spurious responses, we prune the affinity graphs by dynamically detaching redundant and unreliable edges. Extensive experiments on CIFAR-100, ImageNet, and MS-COCO datasets demonstrate the superior performance of the proposed virtual relation matching (VRM) method, where it consistently sets new state-of-the-art records over a range of models, architectures, tasks, and set-ups. 
For instance, VRM for the first time hits 74.0\% accuracy for ResNet50 $\!\to\!$ MobileNetV2 distillation on ImageNet, and improves DeiT-T by 14.44\% on CIFAR-100 with a ResNet56 teacher.
\end{abstract}    
\section{Introduction}
\label{sec:intro}

Deep learning is achieving incredible performance at the cost of increasing model complexity and overheads. As a consequence, large and cumbersome neural models struggle to work in resource-constrained environments. Knowledge distillation (KD)~\cite{kd} has been proposed to address this issue by transferring the knowledge of larger and more capable models to smaller and lightweight ones that are resource-friendly. KD works by minimising the distance between compact representations of knowledge extracted from the teacher and student models. It has found widespread application across a spectrum of downstream tasks~\cite{cirkd, fgfi, fgd, odm3d, unidistill, discofos, ekd, fretal, hfgi3d,  kddlgan, minillm}.

\begin{figure}[t]
\centering
\begin{subfigure}[t]{0.49\linewidth}
  \centering
  \includegraphics[width=\linewidth]{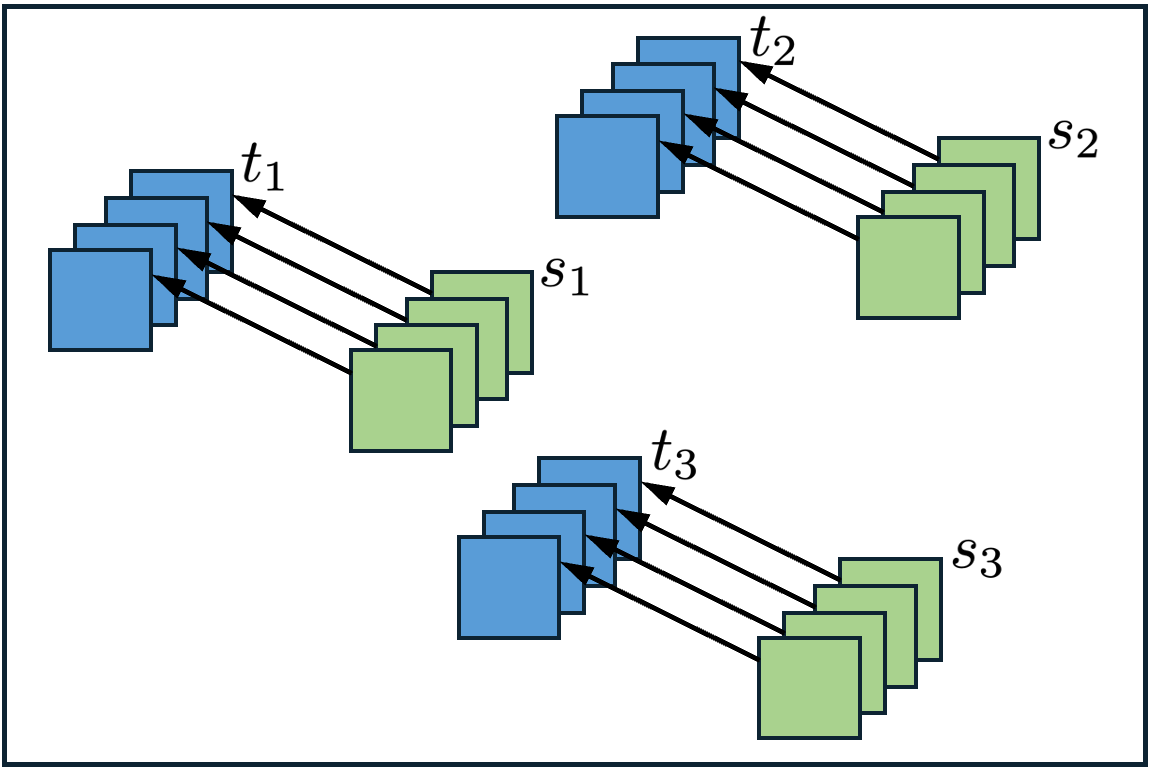}
  \caption{Instance Matching}
  \label{fig:concept_im}
\end{subfigure}
\hfill
\begin{subfigure}[t]{0.49\linewidth}
  \centering
  \includegraphics[width=\linewidth]{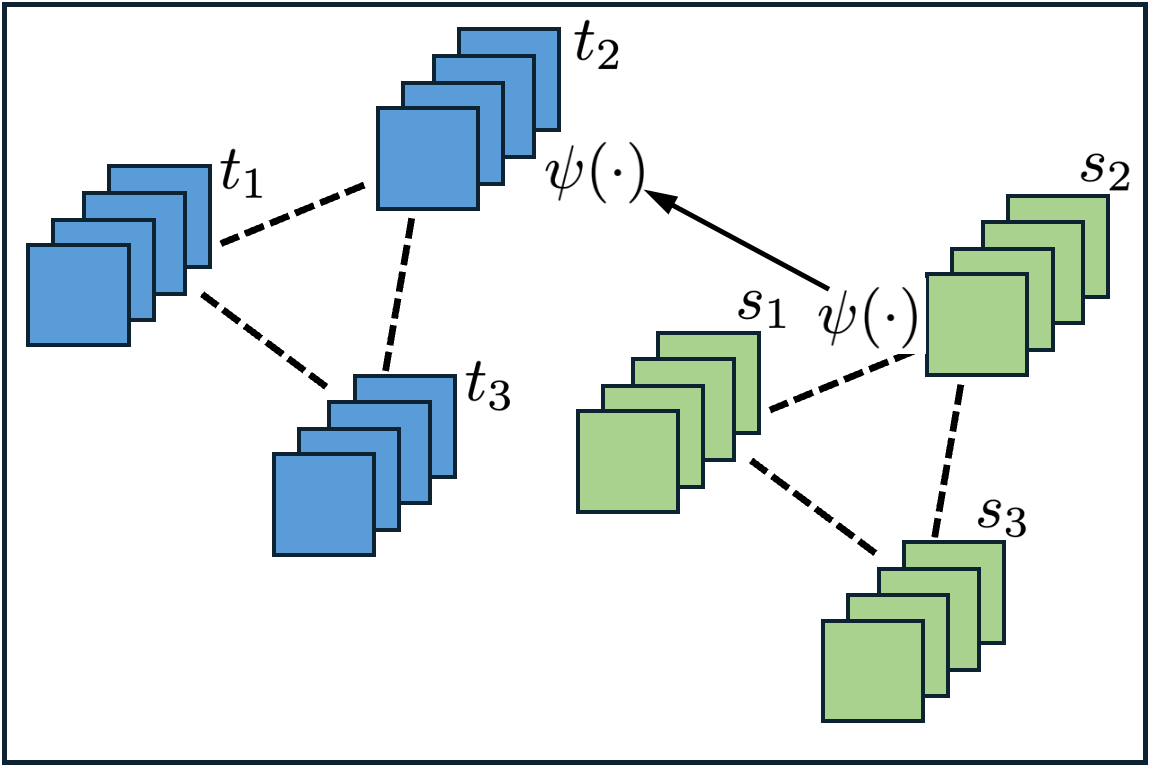}
  \caption{Relation Matching}
  \label{fig:concept_rm}
\end{subfigure}
\\[1mm]
\begin{subfigure}[t]{0.49\linewidth}
  \centering
  \includegraphics[width=\linewidth]{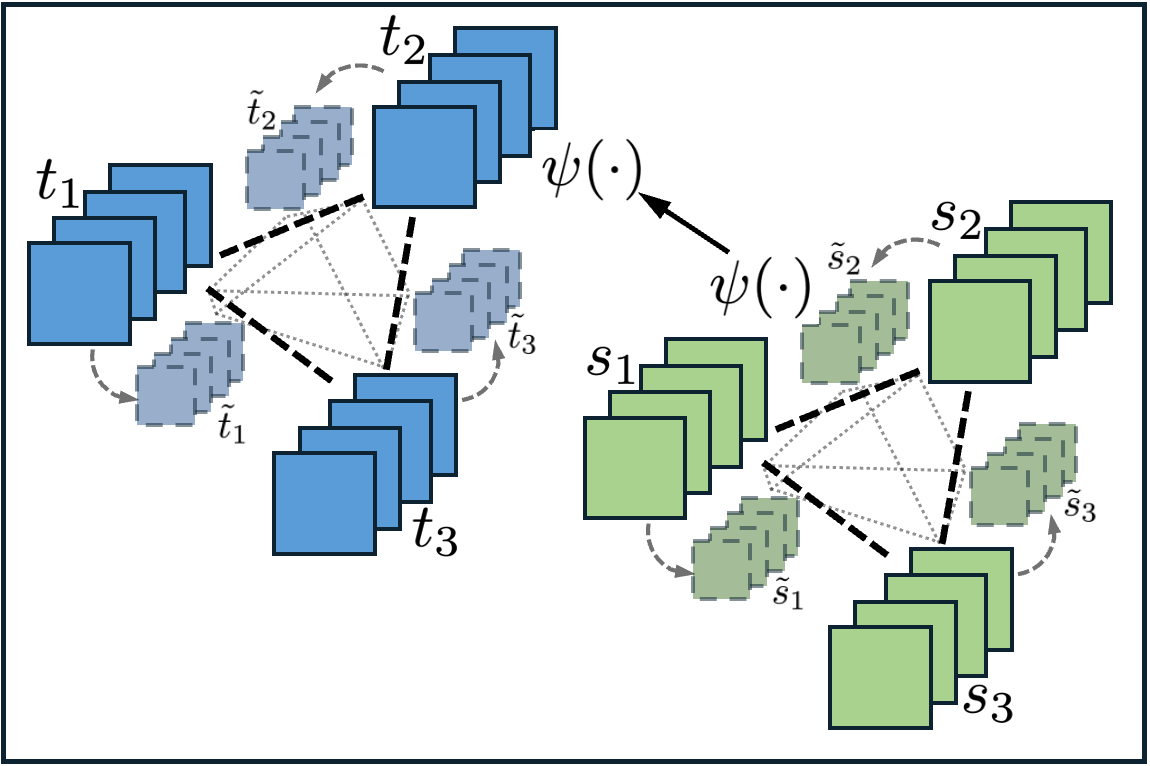}
  \caption{VRM (Dense)}
  \label{fig:concept_vrmd}
\end{subfigure}
\hfill
\begin{subfigure}[t]{0.49\linewidth}
  \centering
  \includegraphics[width=\linewidth]{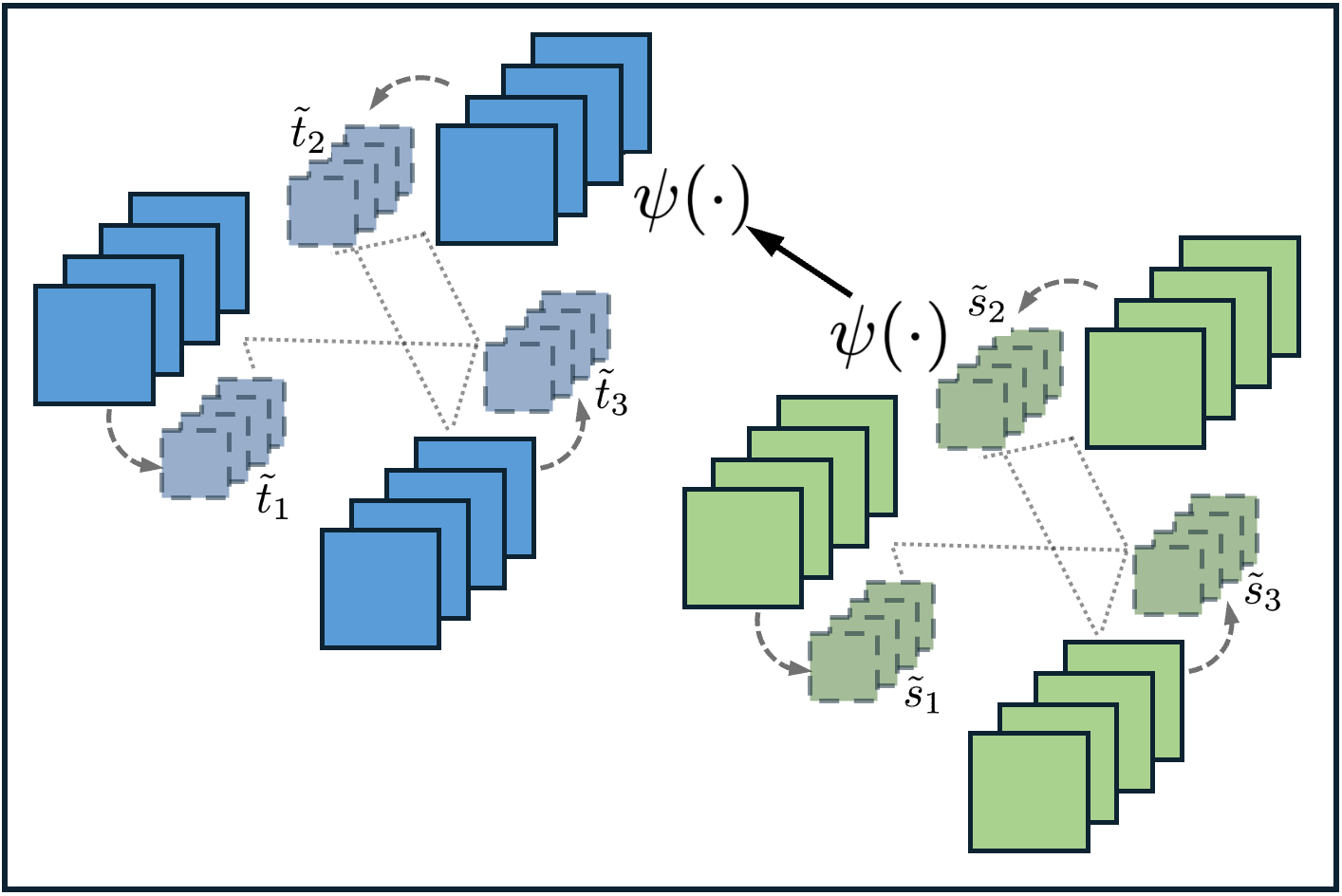}
  \caption{VRM (Pruned)}
  \label{fig:concept_vrms}
\end{subfigure}\vspace{-2mm}
 \caption{Conceptual illustration of VRM compared to existing KD methods based on instance matching and relation matching.} \label{fig:concept} \vspace{-2mm}
\end{figure}

According to the type of knowledge representations to be transferred, KD methods can be broadly categorised into feature-based~\cite{fitnets}, logit-based~\cite{kd}, and relation-based~\cite{rkd} approaches. The former two directly match the feature maps or logit vectors produced by the teacher and student models for each training sample, which is essentially \textit{instance matching} (IM). By contrast, \textit{relation matching} (RM) methods construct and match structured relations extracted within a batch of model responses. A conceptual illustration is presented in Figs.~\ref{fig:concept_im} and~\ref{fig:concept_rm}. 

Instance matching has been the prevailing distillation approach in recent years. Popular KD benchmarks see a dominance by IM methods such as FCFD~\cite{fcfd}, NORM~\cite{norm}, and CRLD~\cite{crld}, with different downstream tasks successfully tackled by directly adopting IM-based distillation~\cite{fgfi, bckd, detrdistill, bevdistill}.
Yet, recent studies discovered that relational knowledge is more robust to variations in neural architectures, data modalities, and tasks~\cite{rkd, spkd}. Meanwhile, methods transferring relations have also achieved promising performance for a range of tasks, including but not limited to segmentation~\cite{cirkd} and detection~\cite{monodistill, stxd}. 

Despite growing interest, relation-based methods still fall significantly short compared to their instance matching counterparts. Even the strongest RM method has been outperformed easily by recent IM solutions~\cite{dist} (see Tabs.~\ref{tab:cifar100} and~\ref{tab:imagenet}).
RM-based methods also struggle with more challenging tasks such as object detection~\cite{dist}. Moreover, previous RM solutions are primarily limited to matching inter-sample~\cite{spkd, pkt, dist}, inter-class~\cite{dist}, or inter-channel~\cite{fsp, ickd} relations via simple Gram matrices. To our best knowledge, no different forms of relations other than these have been proposed since the work of DIST~\cite{dist}.

This paper fills this gap with a new kind of relations for KD -- \textit{inter-view} relations (Fig.~\ref{fig:concept_vrmd}), which seamlessly and compactly integrate with previous inter-sample and inter-class relations. Our designs are motivated by two important observations made about RM methods in a set of pilot experiments: 1) RM methods are more susceptible to overfitting than IM methods; 2) RM methods are subject to an adverse gradient propagation effect. We empirically find that incorporating richer and more diverse relations into the matching objective helps mitigate both issues. 

To this end, we generate \textit{virtual views} of samples through simple transformations, followed by constructing \textit{virtual affinity graphs} and transferring the \textit{virtual relations} between real and virtual samples along the edges. In lieu of Gram matrices that suffer from significant knowledge loss~\cite{spkd, pkt, dist}, we preserve the raw relations along the secondary dimension as auxiliary knowledge which adds to the types and density of relational knowledge transferred. Moreover, we also prune our affinity graphs by striping away both redundant and unreliable edges to further alleviate the propagating gradients of spurious samples (Fig.~\ref{fig:concept_vrms}).

The above insights and remedies altogether lead to a novel \textit{Virtual Relation Matching} (VRM) framework for knowledge distillation. VRM is conceptually simple, easy to implement, and devoid of complicated training procedures.
It is capable of transferring rich, sophisticated knowledge robust to overfitting and spurious signals. VRM sets new state-of-the-art performance for different datasets, tasks, and settings.
Perhaps more significant is that VRM makes relation-based methods regain competitiveness and back in the lead over instance matching approaches in different scenarios. To summarise, the contributions of this work include:
\begin{itemize}
    \item We make an early effort to present comparative analyses of existing KD methods through the lens of training dynamics and sample-wise gradients, and identify overfitting and spurious gradient diffusion as two main cruxes in relational KD methods.
    \item  We distill richer, more diverse relations by generating virtual views, constructing virtual affinity graphs, and matching virtual relations. We also for the first time tackle relational KD with considerations of spurious samples and gradients by pruning redundant and unreliable edges, alongside designs to relax the matching criterion. 
    \item We present the streamlined VRM framework for knowledge distillation, with extensive experimental results on a diversity of models and tasks to highlight its superior performance, alongside rigorous analyses on the soundness and efficiency of our designs. 
\end{itemize}
\begin{figure*}[t] \centering
\begin{subfigure}[t]{0.32\textwidth}
\centering \includegraphics[width=\textwidth, height=0.52\textwidth]{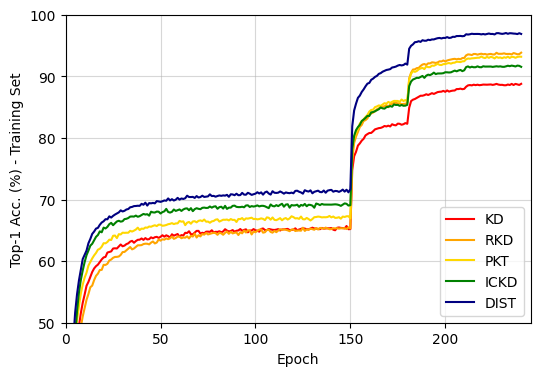} \vspace{-6mm}
\caption{Increasing performance on training set.}
\label{fig:pilot_a1}
\end{subfigure} 
\hfill
\begin{subfigure}[t]{0.32\textwidth}
\centering
\includegraphics[width=\textwidth, height=0.52\textwidth]{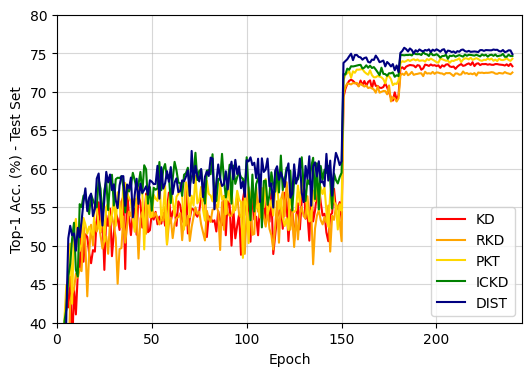} \vspace{-6mm}
\caption{Plateaued performance on validation set.}
\label{fig:pilot_a2}
\end{subfigure}  
\hfill
\begin{subfigure}[t]
{0.35\textwidth}\centering 
\includegraphics[width=\textwidth, height=0.47\textwidth]{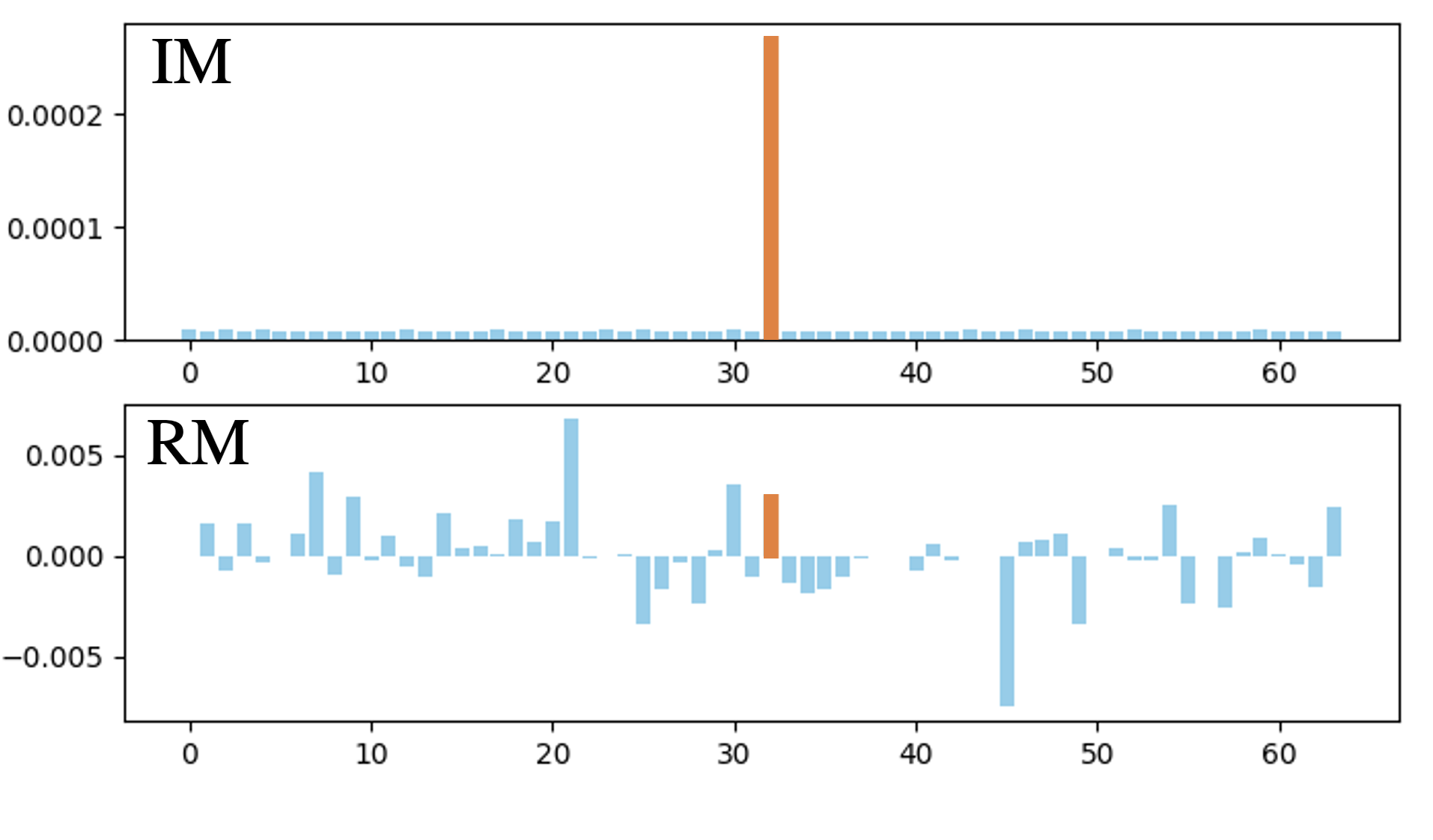}
\vspace{-6mm}
\caption{Diffusion of gradient perturbations within a batch.}
\label{fig:pilot_b}
\end{subfigure} \vspace{-4mm}
\caption{Pilot studies that reveal the overfitting and spurious gradient diffusion issues with RM-based KD.} \vspace{-2mm}
\end{figure*}

\section{Related Work}
\label{sec:related_work}

\noindent \textbf{KD via instance matching.} KD was first proposed in~\citet{kd}, where a student is trained to mimic the prediction of a pre-trained teacher model for each sample. 
Follow-up works have mostly followed such \textit{instance matching} (IM) paradigm, and can be categorised into logit (or prediction)-based and feature-based methods according to what is matched. 
Logit-based KD has evolved from using adaptive-softened logits~\cite{ctkd, mlld} to decoupling target- and non-target logits~\cite{dkd, nkd, ofa} and applying logit transformation~\cite{lskd, ttm}. Others~\cite{takd, dgkd} set up auxiliary ad-hoc networks between teacher and student to facilitate logit transfer.
Feature-based methods minimise the distance between the feature maps~\cite{fitnets, vid, ofd} or salient regions within in features~\cite{at, catkd} from specified layers in both teacher and student networks. 
Some also design sophisticated distillation paths~\cite{semckd, tat}. All these methods are based on instance-wise transfer of knowledge, and are herein referred to as the ``IM'' methods, as illustrated by Fig.~\ref{fig:concept_im}.

\vspace{1.6mm}
\noindent \textbf{KD via relation matching.}
Some works transfer instead mutual relations mined amongst the network outputs extracted from a batch of training instances. These \textit{relation matching} (RM) methods usually involve constructing network outputs into compact relation representations that encode rich higher-order information, as depicted in Fig.~\ref{fig:concept_rm}. Different relation encoding functions mare, including inter-sample~\cite{rkd, irg, pkt, cckd, spkd, dist}, inter-class~\cite{dist}, inter-channel~\cite{fsp, ickd, rsd}, inter-layer~\cite{irg}, and contrastive~\cite{crd} relations. To date, IM solutions have dominated KD with their superior performance, leading top-performing relation-based methods by considerable margins. While many new IM methods are found within the last two years, frustratingly few RM solutions are being proposed. In this work, we strive to close this gap with a new kind of relations for KD --- inter-view virtual relations, and revive relation-based KD by making it overtake its IM counterparts. 

\vspace{1.6mm}
\noindent \textbf{Learning with virtual knowledge.} While not a standalone research topic, learning with virtual knowledge finds relevance in a variety of learning-based problems. For instance, a commonly used paradigm in 3D vision tasks is to learn (or construct) from the raw data a virtual view or representation as auxiliary knowledge in solving the main task, ranging from object reconstruction~\cite{vvn} and optical flow~\cite{aleotti2021} to 3D semantic segmentation~\cite{vmf}, monocular 3D object detection~\cite{pseudostereo}, and 3D GAN inversion~\cite{hfgi3d}.
More broadly, many data-efficient learning methods also share the spirit of utilising virtual knowledge. For instance, self-supervised learning methods generate virtual views of the unlabelled data to enable the learning of pretext tasks~\cite{rotnet, simclr}. Another popular paradigm is transformation-invariant representation learning~\cite{pirl, fixmatch} in semi-supervised learning and domain adaptation. It enforces consistency between representations learnt for a raw sample and a virtual view of it. The virtual view is often obtained by applying semantic-preserving transformations to the raw sample~\cite{randaugment, autoaugment}. This work is more related to this later paradigm, but involves learning with virtual knowledge in a different context, via different approaches, and for a different problem.
\section{Preliminaries}
\label{sec:preliminaries}
KD methods generally employ a cross-entropy (CE) loss and a distillation loss to supervise student learning. The CE loss is computed between student logits $\mathbf{z}^{s}_i$ for each sample and its ground-truth label $\mathbf{y}_i$.
The distillation loss matches teacher and student outputs via a distance metric $\phi(\cdot)$. In vanilla KD~\cite{kd}, $\phi(\cdot)$ is the Kullback–Leibler divergence (KLD) between teacher logits $\mathbf{z}^{t}$ and student logits $\mathbf{z}^{s}$: \vspace{-1mm}
\begin{align}
    \mathcal{L}^{\text{KD}}_i &= 
    \phi_{\text{KLD}}(\mathbf{z}^s_{i}, \mathbf{z}^t_{i}) \nonumber \\ &= \tau^2\sum_{j=1}^C \sigma_j(\mathbf{z}^{t}_{i}/\tau)\log\frac{\sigma_j(\mathbf{z}^{t}_{i}/\tau)}{\sigma_j(\mathbf{z}^{s}_{i}/\tau)}, \vspace{-3mm}
\end{align}
\noindent where $\sigma(\cdot)$ is the Softmax operation with temperature parameter $\tau$, and $C$ is the number of classes. For feature-based methods, $\phi(\cdot)$ can be the mean squared error (MSE) between teacher and student feature maps for each instance~\cite{fitnets}, \textit{i.e.}, $  \mathcal{L}^{\text{KD}}_i =
    \phi_{\text{MSE}}(\mathbf{f}^s_{i}, \mathbf{f}^t_{i}) $. 
    
For relation-based methods, a relation encoding function $\psi(\cdot)$ first abstracts teacher or student outputs of all instances within a training batch into a relational representation, before applying $\phi(\cdot)$ to match these relational representations between the teacher and the student. For example, the KD objectives of DIST~\cite{dist} take the general form of:
\begin{align}
    \mathcal{L}^{\text{KD}} =
    \phi(\psi(\mathbf{z}^s_{1}, \mathbf{z}^s_{2}, ..., \mathbf{z}^s_{B}), \psi(\mathbf{z}^t_{1}, \mathbf{z}^t_{2}, ..., \mathbf{z}^t_{B})),
\end{align}
\noindent where $B$ is the batch size, and subscript $i$ of $ \mathcal{L}^{\text{KD}}$ is dropped as the loss is computed for a batch of instances. DIST uses both inter-class and inter-sample relation encoders as $\psi(\cdot)$.

\section{Pilot Studies} \label{sec:pilot_expts}
\noindent \textbf{Training dynamics of KD methods.}
We examine the training dynamics of different relation-based methods on CIFAR-100 with  ResNet32$\times4$ $\!\to\!$ ResNet8$\times4$ as the teacher-student pair. In Figs.~\ref{fig:pilot_a1} and ~\ref{fig:pilot_a2}, we immediately notice that relational methods achieve significantly higher training accuracy, but they only marginally lead or even fall short in test accuracy compared to IM-based KD~\cite{kd}. We hypothesise that relational methods are more prone to overfitting. This is expected given that the optimality of IM matching (cond. $\mathcal{A}$) implies the optimality of relation matching (cond. $\mathcal{B}$), while the converse does not hold. Concretely, $\mathcal{A} \Rightarrow \mathcal{B} \; \land \; \neg \mathcal{B} \Rightarrow \neg \mathcal{A}$. In other words, relation matching is a weaker and less constrained objective than IM, which makes the student more readily fit the teaching signals and not generalise well. Thus, we conclude that \textit{C1: relation matching methods are more prone to overfitting.} 

\vspace{1.6mm}
\noindent \textbf{Gradient analysis of KD methods.} 
We investigate the gradient patterns within a batch when a spurious sample produces a major misguiding signal. To this end, we first generate two random vectors $\mathbf{x}, \mathbf{y} \sim \mathcal{N}(0, 1) \text{ for } \mathbf{x}, \mathbf{y} \in \mathbb{R}^{B \times D}$, where $B$ is the batch size and $D$ is the dimension of per-sample predictions, $\mathbf{x}$ is taken as the sample-wise predictions, and $\mathbf{y}$ the supervision signals. We then add a noise vector $\epsilon = c \cdot  \mathbf{z}$ to $\mathbf{x}_t$, where $\mathbf{z} \sim \mathcal{N}(0, 1) $ and $c$ is a scaling factor. In this case, $\mathbf{x}_t + \epsilon$ becomes a spurious prediction within our batch. We compute the loss from $\mathbf{x}$ and $\mathbf{y}$ using either IM or RM objectives, and consider the change in sample-wise gradients $\mathbf{g}$ within the batch upon the injection of the spurious sample. Formally, we visualise:
\begin{align}
     \Delta \mathbf{g} &= \left[ \|\frac{\partial \mathcal{L}}{\partial (\mathbf{x'}_i)}\|_2 - \|\frac{\partial \mathcal{L}}{\partial (\mathbf{x}_i)}\|_2 \right]_{i=1}^{B}, \quad \nonumber
\\ \text{s. t. } \mathbf{x'}_i &= \mathbf{x}_i + \epsilon \cdot \mathbb{I}(i=t), \quad \mathcal{L} \in \{ \mathcal{L}_{IM}, \mathcal{L}_{RM} \}.
\end{align}

\begin{figure*}[t] \centering
\begin{minipage}[t]{0.63\textwidth} \centering 
\begin{subfigure}[t]{0.49\textwidth}
\centering \includegraphics[width=\textwidth, height=0.47\textwidth]{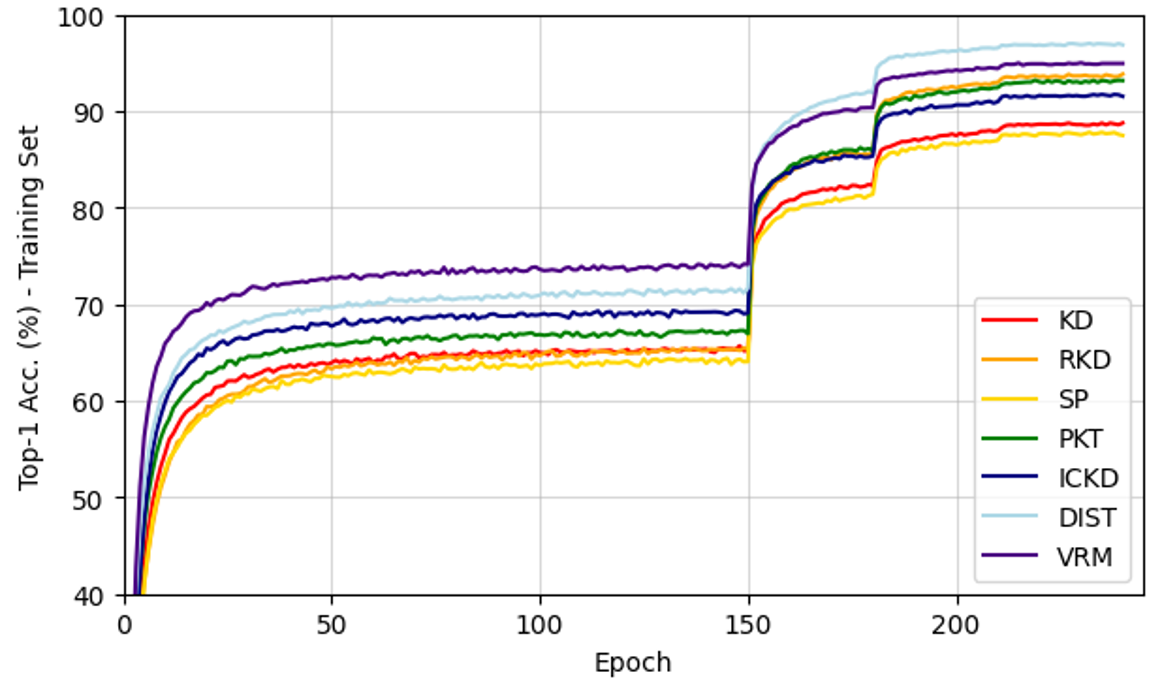} \vspace{-5mm}
\captionsetup{font=scriptsize}
\caption{Training accuracy} \label{fig:train_dynamics}
\end{subfigure}
\hfill
\begin{subfigure}[t]{0.49\textwidth}
\centering \includegraphics[width=\textwidth, height=0.47\textwidth]{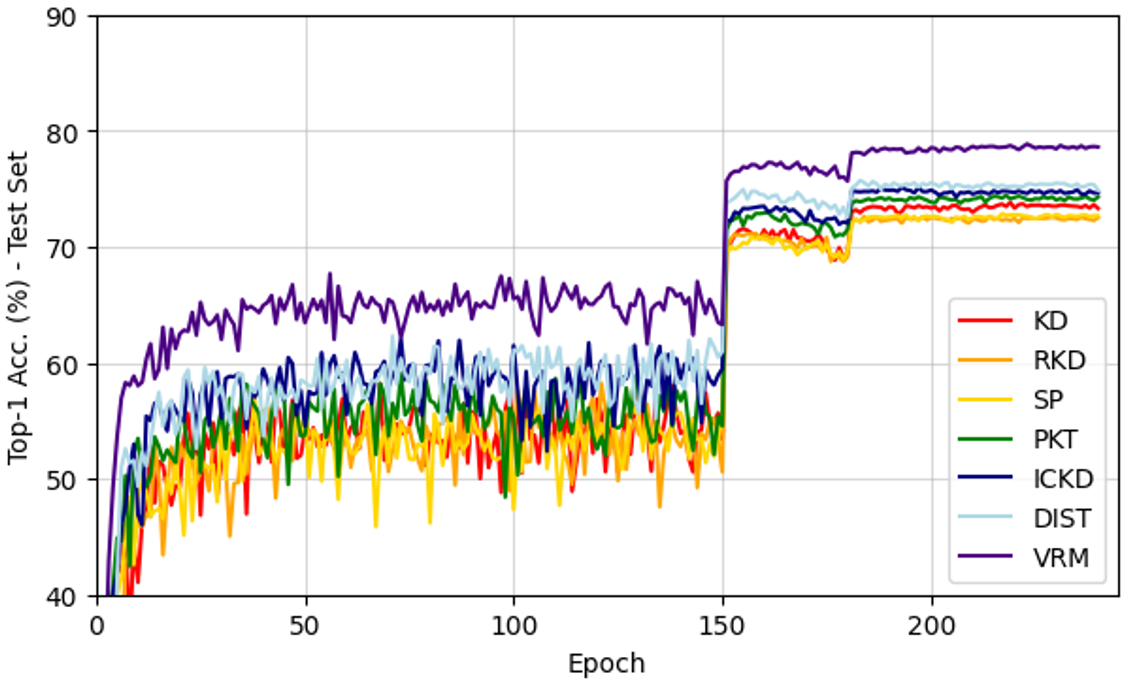} \vspace{-5mm}
\captionsetup{font=scriptsize}
\caption{Test accuracy} \label{fig:val_dynamics}
\end{subfigure} \vspace{-3.5mm}
\caption{Alleviated overfitting and improved training dynamics.}
\end{minipage} \hfill
\begin{minipage}[t]{0.36\textwidth}
\vspace{-25mm}
\includegraphics[width=\textwidth, height=0.42\textwidth]{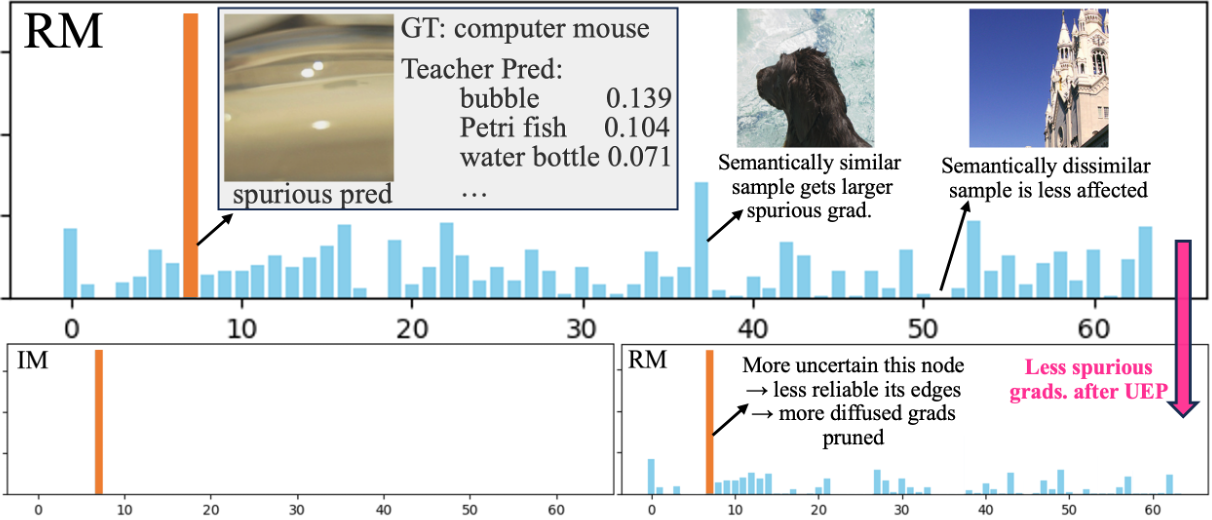} \vspace{-6mm}
\caption{Alleviated spurious gradients.} \label{fig:pilot_in}
\end{minipage}  \vspace{-3mm}
\end{figure*}

In Fig.~\ref{fig:pilot_b} ($B=64$ and $t=32$), when the IM objective is used, only the spurious sample receives a prominent gradient. Whereas for an RM objective, many other samples receive significant gradients as they are directly connected to $\mathbf{x}_t$ in the computational graph of the RM loss. In other words, the spurious signals produced by one malign prediction will propagate to and affect all samples within a batch (in fact, those closer to $\mathbf{x}_t$ within the prediction manifold are more strongly affected). This means other sample-wise predictions will be significantly updated only to accommodate a malign prediction, even if they are already in relatively good shape. Through this  investigation, we discover that \textit{C2: relation matching methods are more prone to the adverse impact of spurious samples}. We also present this pilot analysis on ImageNet training in Fig.~\ref{fig:pilot_in}.

For \textit{C1}, common approaches to combat overfitting include the incorporation of richer learning signals and regularisation, which for RM-based KD methods means richer relations constructed and transferred. For \textit{C2}, an intuitive solution is to identify and suppress the effect of spurious predictions or relations or to slacken the matching criterion. Guided by these principles, we now proceed to formally build up our method step by step.

\begin{table*}[t] \scriptsize 
\centering
\resizebox{0.90\textwidth}{!}{
\setlength{\tabcolsep}{1pt} \vspace{-3mm}
\begin{tabular}{cc|ccccc|cccc}
\toprule
\multirow{2}{*}{\shortstack{Teacher\\Student}} & \multirow{4}{*}{Venue} & \multirow{2}{*}{\shortstack{ResNet56\\ResNet20}}  &  \multirow{2}{*}{\shortstack{ResNet32$\times$4\\ResNet8$\times$4}} & \multirow{2}{*}{\shortstack{WRN-40-2\\WRN-16-2}} & \multirow{2}{*}{\shortstack{WRN-40-2\\WRN-40-1}}  & \multirow{2}{*}{\shortstack{VGG13\\VGG8}} & \multirow{2}{*}{\shortstack{ResNet32$\times$4\\ShuffleNetV2}} & \multirow{2}{*}{\shortstack{VGG13\\MobileNetV2}} & \multirow{2}{*}{\shortstack{ResNet50\\MobileNetV2}} & \multirow{2}{*}{\shortstack{WRN-40-2\\ShuffleNetV1}} \vspace{-1pt} \\
& & & & & & & & & & \\
\cmidrule(lr){1-1} \cmidrule(lr){3-11}
Teacher & & 72.34 & 79.42 & 75.61 & 75.61 & 74.64 & 79.42 & 74.64 & 79.34 & 75.61 \vspace{-0.5pt} \\
Student & & 69.06 & 72.50 & 73.26 & 71.98 & 70.36 & 71.82 & 64.60 & 64.60 & 70.50 \vspace{-0.5pt} \\
\midrule
\multicolumn{11}{c}{\textit{Feature-based}} \\ 
FitNets~\cite{fitnets} & ICLR'15 & 69.21 & 73.50 & 73.58 & 72.24 & 71.02 & 73.54 & 64.16 & 63.16 & 73.73 \vspace{-0.5pt} \\
AT~\cite{at} & ICLR'17 & 70.55 & 73.44 & 74.08 & 72.77 & 71.43 & 72.73 & 59.40 & 58.58 & 73.32 \vspace{-0.5pt}  \\
CRD~\cite{crd} & ICLR'20 & 71.16 & 75.51 & 75.48 & 74.14 & 73.94 & 75.65 & 69.63 & 69.11 & 76.05 \vspace{-0.5pt}  \\
SRRL~\cite{srrl} & ICLR'21 & 71.13 & 75.33 & 75.59 & 74.18 & 73.44 & - & - & - & - \vspace{-0.5pt}  \\
PEFD~\cite{pefd} & NeurIPS'22 & 70.07 & 76.08 & 76.02 & 74.92 & 74.35 & - & - & - & - \vspace{-0.5pt} \\
TaT~\cite{tat} & CVPR'22 & 71.59 & 75.89 & 76.06 & 74.97 & 74.39 & - & - & - & - \vspace{-0.5pt} \\
ReviewKD~\cite{reviewkd} & CVPR'21 & 71.89 & 75.63 & 76.12 & 75.09 & 74.84 & 77.78 & 70.37 & 69.89 & 77.14 \vspace{-0.5pt} \\
NORM~\cite{norm} & ICLR'23 & 71.35 & 76.49 & 75.65 & 74.82 & 73.95 & 78.32 & 69.38 & 71.17 & 77.63 \vspace{-0.5pt} \\
FCFD~\cite{fcfd} & ICLR'23 & 71.96 & 76.62 & 76.43 & 75.46 & 75.22 & 78.18 & 70.65 & 71.00 & 77.99 \vspace{-0.5pt} \\
\midrule
\multicolumn{11}{c}{\textit{Logit-based}} \\
KD~\cite{kd} & arXiv'15 & 70.66 & 73.33 & 74.92 & 73.54 & 72.98 & 74.45 & 67.37 & 67.35 & 74.83 \vspace{-0.5pt} \\
TAKD~\cite{takd} & AAAI'20 & 70.83 & 73.81 & 75.12 & 73.78 & 73.23 & 74.82 & 67.91 & 68.02 & 75.34 \vspace{-0.5pt} \\
CTKD~\cite{ctkd} & AAAI'23 & 71.19 & 73.79 & 75.45 & 73.93 & 73.52 & 75.31 & 68.46 & 68.47 & 75.78 \vspace{-0.5pt} \\
NKD~\cite{nkd} & ICCV'23 & 70.40 & 76.35 & 75.24 & 74.07 & 74.86 & 76.26 & 70.22 & 70.76 & 75.96 \vspace{-0.5pt} \\
DKD~\cite{dkd} & CVPR'22 & 71.97 & 76.32 & 76.24 & 74.81 & 74.68 & 77.07 & 69.71 & 70.35 & 76.70 \vspace{-0.5pt} \\
LSKD~\cite{lskd} & CVPR'24 & 71.43 & 76.62 & 76.11 & 74.37 & 74.36 & 75.56 & 68.61 & 69.02 & - \vspace{-0.5pt} \\
TTM~\cite{ttm} & ICLR'24 & 71.83 & 76.17 & 76.23 & 74.32 & 74.33 & 76.55 & 69.16 & 69.59 & 75.42 \vspace{-0.5pt} \\
CRLD~\cite{crld} & MM'24 & 72.10 & 77.60 & 76.45 & 75.58 & 75.27 & 78.27 & 70.39 & 71.36 & -\vspace{-0.5pt} \\
\midrule
\multicolumn{11}{c}{\textit{Relation-based}} \\
RKD~\cite{rkd} & CVPR'19 & 69.61 & 71.90 & 73.35 & 72.22 & 71.48 & 73.21 & 64.52 & 64.43 & 72.21 \vspace{-0.5pt} \\
CC~\cite{cckd} & CVPR'19 & 69.63 & 72.97 & 73.56 & 72.21 & 70.71 & 71.29 & 64.86 & 65.43 & 71.38 \vspace{-0.5pt} \\
SP~\cite{spkd} & ICCV'19 & 69.67 & 72.94 & 73.83 & 72.43 & 72.68 & 74.56 & 66.30 & 68.08 & 74.52 \vspace{-0.5pt} \\
ICKD~\cite{ickd} & ICCV'21 & 71.76 & 75.25 & 75.64 & 74.33 & 73.42 & - & - & - & - \vspace{-0.5pt} \\
DIST~\cite{dist}$\dagger$ & NeurIPS'22 & 71.75 & 76.31 & - & 74.73 & - & 77.35 & 68.50 & 68.66 & 76.40 \vspace{-0.5pt} \\ 
\cellcolor{sgray} \textbf{VRM} & \cellcolor{sgray} - & \cellcolor{sgray} \textbf{72.09} & \cellcolor{sgray} \textbf{78.76} & \cellcolor{sgray} \textbf{77.47} & \cellcolor{sgray} \textbf{76.46} & \cellcolor{sgray} \textbf{76.19} & \cellcolor{sgray} \textbf{79.34} & \cellcolor{sgray} \textbf{71.66} & \cellcolor{sgray} \textbf{72.30} & \cellcolor{sgray} \textbf{78.62} \\
\bottomrule
\end{tabular}} \vspace{-1mm}
\caption{Results for same- and different-model teacher-student pairs on CIFAR-100. $\dagger$: using re-trained, stronger teachers.}
\label{tab:cifar100}
\vspace{-2mm}
\end{table*}
\section{Method}
\label{sec:method}

\subsection{Constructing Inter-Sample Relations}

We first construct relation graph $\mathcal{G}^{IS}$ that encodes inter-sample affinity within a batch of sample predictions $\{ \mathbf{z}_i \}^B_{i=1}$. 
Different from ~\cite{rkd, spkd}, our relations are constructed from predicted logits which embed more compact categorical knowledge. We use the pairwise distance between instance-wise predictions within a batch as our measure of affinity. Existing methods leverage the Gram matrices~\cite{cckd, pkt, spkd, dist} to encode inter-sample relations, but we find that this leads to collapsed inter-class knowledge via the inner product operation. Instead, our pairwise distance preserves the inter-class knowledge along the secondary dimension (\textit{i.e.}, the class dimension), which enables such information to be explicitly transferred as auxiliary knowledge alongside the matching of inter-sample relations.

Thus, we have constructed a dense relation graph $\mathcal{G}^{IS}$, which comprises $B$ vertices and $B \times B$ edges: $\mathcal{G}^{IS}=(\mathcal{V}^{IS}, \mathcal{E}^{IS})$.  Each vertex in $\mathcal{G}^{IS}$ represents the prediction vector $\mathbf{z} \in \mathbb{R}^{C}$ for one instance within a batch, and is connected to all instances within the batch including itself. The attribute of edge $\mathcal{E}^{IS}_{i,j}$ connecting vertices $i$ and $j$ describes the class-wise relations between the predictions of instances $i$ and $j$. In practice, we can organise all edges into matrix $\mathcal{E}^{IS} \in \mathbb{R}^{B\times B \times C}$, in which:
\begin{align}
    \mathcal{E}^{IS}_{i,j} = \frac{\mathbf{z}_i - \mathbf{z}_j}{\|\mathbf{z}_i - \mathbf{z}_j\|_2} \in \mathbb{R}^C \quad \text{for} \quad i, j \in [1, B],
\end{align}
where we empirically find that normalisation along the secondary dimension helps regularise relations and improves performance. Concretely, $\mathcal{E}$ encodes inter-sample class-wise relations within a batch of $B$ training samples. This design is different from and empirically more effective than previous relation formulations.

\subsection{Constructing Inter-Class Relations}
Although $\mathcal{G}^{IS}$ encodes rich inter-sample relations within a batch, it fails to explicitly model inter-class correlation patterns that are also beneficial structured knowledge~\cite{dist}. Therefore, we propose to build a novel inter-class batch-wise relation graph $\mathcal{G}^{IC} = (\mathcal{V}^{IC}, \mathcal{E}^{IC})$, instead of the Gram matrices in the inner product space as in~\cite{dist}. The construction of $\mathcal{G}^{IC}$ mirrors that of $\mathcal{G}^{IS}$. Vertices in $\mathcal{G}^{IC}$ are the class-wise logit vectors $\mathbf{w} \in \mathbb{R}^{B} $. Each edge in $\mathcal{E}^{IC} \in \mathbb{R}^{C \times C \times B}$ embeds the pairwise difference between the $i$-th and $j$-th per-class vectors. Concretely:
\begin{align}
    \mathcal{E}^{IC}_{i,j} = \frac{\mathbf{w}_i - \mathbf{w}_j}{\|\mathbf{w}_i - \mathbf{w}_j\|_2} \in \mathbb{R}^B \quad \text{for} \quad i, j \in [1, C].
\end{align}

Our inter-class relations preserve the batch-wise discrepancies by treating them as a dimension of additional knowledge (reciprocal to the case of inter-sample relations), which is unlike any other previous methods~\cite{dist}. We will demonstrate in Tab.~\ref{tab:abl_psi} that our novel formulation of inter-sample and inter-class relations by preserving the raw affinity knowledge along the secondary dimension perform significantly better than previous relation encoders $\psi(\cdot)$ via Gram matrices~\cite{rkd, pkt, spkd, dist} or third-order angular distances~\cite{rkd}.

\subsection{Constructing Virtual Relations}
\label{sec:virtual_relation}
For each prediction $\mathbf{z}_i$ within a batch $\{ \mathbf{z}_i \}^B_{i=1}$, we create a virtual view of it, denoted as ``$\Tilde{\mathbf{z}}_i$'', by applying semantic-preserving transformations to original image $\mathbf{x}_i$. While other transformations are applicable, we choose RandAugment~\cite{randaugment} that applies stochastic image transformations (see Supplementary Material for details). With our batch of predictions augmented into $\{ \mathbf{z}_i, \Tilde{\mathbf{z}}_i \}^B_{i=1}$, we can construct a larger inter-sample edge matrix $\mathcal{E}^{IS} \in \mathbb{R}^{2B \times 2B \times C}$ and a larger inter-class edge matrix $\mathcal{E}^{IC} \in \mathbb{R}^{C \times C \times 2B}$. 

From the perspective of sample views, our new $\mathcal{E}^{IS}$ and $\mathcal{E}^{IC}$ constructed from a batch of real and virtual samples essentially capture three types of knowledge, namely relations amongst real views (denoted as ``real-real''), relations amongst virtual views (``virtual-virtual''), and relations between pairs of real and virtual views (``real-virtual''). For instance, a real-virtual edge that connects real vertex $m$ and virtual vertex $n$ in $\mathcal{E}^{IS}$ is computed as:
\begin{equation}
    \mathcal{E}^{IS}_{m,n} = \frac{\mathbf{z}_m - \tilde{\mathbf{z}}_n}{\|\mathbf{z}_m - \tilde{\mathbf{z}}_n \|_2} \in \mathbb{R}^C.
\end{equation}

\subsection{Pruning into Sparse Graphs} \label{sec:method_pruning}
\noindent \textbf{Pruning redundant edges.}
The augmented $\mathcal{E}^{IS}$ in Sec.~\ref{sec:virtual_relation} contains $2B \times 2B$ edges, leading to quadrupled overheads. For better efficiency, we prune $\mathcal{G}^{IS}$ into sparse graphs. Noticing that $\mathcal{E}^{IS}$ is symmetric along its diagonals, we first prune its redundant half to save up to $50\%$ edge count. We also remove intra-view edges to get a further $50\%$ reduction, as we empirically find them redundant and harm knowledge transfer. 
For $\mathcal{G}^{IC}$, we decompose the augmented batch of predictions of size $2B$ into a real-view batch and a virtual-view batch, each of size $B$, and in lieu use the inter-sample batch-wise affinity vectors between them as its vertices. Compared to their original intra-view formulation of size $2B$ in Sec.~\ref{sec:virtual_relation}, this  design encodes purely inter-view affinity knowledge with halved parameters. With redundant edge pruning (REP), our graphs now become sparse and the remaining edges can again be rearranged into compact matrices: $\mathcal{E}^{ISV} \in \mathbb{R}^{B \times B \times C}$ and $\mathcal{E}^{ICV} \in \mathbb{R}^{C \times C \times B}$.

\vspace{1.6mm}
\noindent \textbf{Pruning unreliable edges.}
To mitigate the diffusive effect of spurious predictions discovered earlier, we further identify and prune the unreliable edges. In previous graph learning works, the absolute certainty of two vertices are often used to determine the reliability of an edge. For instance, REM~\cite{rem} computes the reliability of an edge as the mean of the maximum predicted probabilities of two samples (\textit{i.e.}, two vertices). However, we argue this could bias the learning towards easy samples. Instead, we measure the discrepancy between two predictions. The larger this discrepancy, the more unreliable the relation constructed.  The unreliable edge pruning (UEP) criterion is given by $\mathcal{E}^{ISV}_{i,j} = \emptyset \quad \text{if} \quad \text{H}(\mathbf{z}^s_i,\tilde{\mathbf{z}^s_j}) > P_{m}$
where $\text{H}(\cdot)$ computes the joint entropy (JE) between two predictions and $P_m$ is the $m$-th percentile within the batch. 
Note that the criterion is enforced on student predictions, resulting in adaptive and dynamic pruning as different edges get pruned in each iteration, which improves learning most of the time.

\begin{table*}[t] \centering
\begin{minipage}{0.49\textwidth} \centering
\renewcommand{\arraystretch}{0.92}
\resizebox{0.80\textwidth}{!}{
\begin{tabular}{cccc} \toprule
\setlength{\tabcolsep}{2pt}
\multirow{2}{*}{\shortstack{Teacher\\Student}} & \multirow{4}{*}{Venue} &  \multirow{2}{*}{\shortstack{ResNet34\\ResNet18}} & \multirow{2}{*}{\shortstack{ResNet50\\MobileNetV1}}  \vspace{-1mm} \\
& & & \\
\cmidrule(lr){1-1} \cmidrule(lr){3-4}
Teacher & & 73.31/91.42 & 76.16/92.86 \\
Student & & 69.75/89.07 & 68.87/88.76 \vspace{-1mm} \\
\midrule
\multicolumn{4}{c}{\textit{Feature-based}} \\
AT~\cite{at} & ICLR'17 & 70.69/90.01 & 69.56/89.33 \\
OFD~\cite{ofd} & ICCV'19 & 70.81/89.98 & 71.25/90.34 \\
CRD~\cite{crd} & ICLR'20 & 71.17/90.13 & 71.37/90.41 \\
CAT-KD~\cite{catkd} & CVPR'23 & 71.26/90.45 & 72.24/91.13 \\
SimKD~\cite{simkd} & CVPR'22 & 71.59/90.48 & 72.25/90.86 \\
ReviewKD~\cite{reviewkd} & CVPR'21 & 71.61/90.51 & 72.56/91.00 \\
SRRL~\cite{srrl} & ICLR'21 & 71.73/90.60 & 72.49/90.92 \\
PEFD~\cite{pefd} & NeurIPS'22 & 71.94/90.68 & 73.16/91.24 \\
RSD~\cite{rsd} & ICCV'25 & 72.18/90.75 & 73.05/91.26 \\
FCFD~\cite{fcfd} & ICLR'23 & 72.24/90.74 & 73.37/91.35 \\
\midrule
\multicolumn{4}{c}{\textit{Logit-based}} \\
KD~\cite{kd} & arXiv'15 & 70.66/89.88 & 68.58/88.98 \\
TAKD~\cite{takd} & AAAI'20 & 70.78/90.16 & 70.82/90.01 \\
DKD~\cite{dkd} & CVPR'22 & 71.70/90.41 & 72.05/91.05 \\
SDD~\cite{sdd} & CVPR'24 & 71.14/90.05 & 72.24/90.71 \\
TTM~\cite{ttm} & ICLR'24 & 72.19/- & 73.09/- \\
LSKD~\cite{lskd} & CVPR'24 & 72.08/90.74 & 73.22/91.59 \\
CRLD~\cite{crld} & MM'24 & 72.37/90.76 & 73.53/91.43 \\
\midrule
\multicolumn{4}{c}{\textit{Relation-based}} \\
RKD~\cite{rkd} & CVPR'19 & 71.34/90.37 & 71.32/90.62 \\
CC~\cite{cckd} & CVPR'19 & 70.74/- & - \\ 
ICKD~\cite{ickd} & ICCV'21 & 72.19/90.72 & - \\
DIST~\cite{dist} & NeurIPS'22 & 72.07/90.42 & 73.24/91.12 \\
\cellcolor{sgray} \textbf{VRM} & \cellcolor{sgray} - & \cellcolor{sgray} \textbf{72.67}/\textbf{90.68} & \cellcolor{sgray} \textbf{74.16}/\textbf{91.78} \vspace{-0.5pt} \\
\bottomrule
\end{tabular}} \vspace{-1mm}
\caption{Results on ImageNet.} \label{tab:imagenet}
\end{minipage}%
\begin{minipage}{0.49\textwidth} \centering
\vspace{0mm}
\resizebox{0.91\textwidth}{!}{
\begin{tabular}{ccccccc} \toprule
 & \multicolumn{3}{c}{ResNet101 $\!\to\!$ ResNet18} & \multicolumn{3}{c}{ResNet50 $\!\to\!$ MobileNetV2} \\
 & $\text{AP}_{}$ & $\text{AP}_{50}$ & $\text{AP}_{75}$ & $\text{AP}_{}$ & $\text{AP}_{50}$ & $\text{AP}_{75}$ \\
\multirow{2}{*}{\shortstack{Teacher\\[2pt]Student}}  &  \multirow{2}{*}{\shortstack{42.04\\[2pt]33.26}} &  \multirow{2}{*}{\shortstack{62.48\\[2pt]53.61}} &  \multirow{2}{*}{\shortstack{45.88\\[2pt]35.26}}  &  \multirow{2}{*}{\shortstack{40.22\\[2pt]29.47}}  &  \multirow{2}{*}{\shortstack{61.02\\[2pt]48.87}} &  \multirow{2}{*}{\shortstack{43.81\\[2pt]30.90}} \\
& & & & & & \\ \midrule
\multicolumn{7}{c}{\textit{Feature-based}} \\
FitNets~\cite{fitnets} & 34.43 & 54.16 & 36.71 & 30.20 & 49.80 & 31.69 \vspace{-0.5pt} \\
FGFI~\cite{fgfi} & 35.44 & 55.51 & 38.17 & 31.16 & 50.68 & 32.92 \vspace{-0.5pt} \\
TAKD~\cite{takd} & 34.59 & 55.35 & 37.12 &     31.26 & 51.03 & 33.46 \vspace{-0.5pt} \\
ReviewKD~\cite{reviewkd} & 36.75 & 56.72 & 34.00 & 33.71 & 53.15 & 36.13 \vspace{-0.5pt} \\
FCFD~\cite{fcfd} & 37.37 & 57.60 & 40.34 & 34.97 & 55.04 & 37.51 \vspace{-0.5pt} \\
\multicolumn{7}{c}{\textit{Logit-based}} \\
KD~\cite{kd} & 33.97 & 54.66 & 36.62 & 30.13 & 50.28 & 31.35 \vspace{-0.5pt} \\
CTKD~\cite{ctkd} & 34.56 & 55.43 & 36.91 & 31.39 & 52.34 & 33.10 \vspace{-0.5pt} \\
LSKD~\cite{lskd} & - & - & - & 31.74 & 52.77 & 33.40 \vspace{-0.5pt} \\
DKD~\cite{dkd} & 35.05 & 56.60 & 37.54 & 32.34 & 53.77 & 34.01 \vspace{-0.5pt} \\
\multicolumn{7}{c}{\textit{Relation-based}} \\
\cellcolor{sgray} \textbf{VRM} & \cellcolor{sgray} \textbf{35.46} & \cellcolor{sgray} \textbf{56.91} & \cellcolor{sgray} \textbf{37.93} & \cellcolor{sgray} \textbf{32.67} & \cellcolor{sgray} \textbf{53.96} & \cellcolor{sgray} \textbf{34.48} \vspace{-0.5pt} \\
\bottomrule
\end{tabular}}
\vspace{-1.0mm}
\caption{Results on MS-COCO. All detectors are based on a Faster-RCNN~\cite{fasterrcnn} backbone.} \label{tab:coco} 
\vspace{4.5mm}
\renewcommand{\arraystretch}{1.0}
\resizebox{\textwidth}{!}{
\setlength{\tabcolsep}{1.3pt}
\begin{tabular}{cccccccccccc}
\toprule
Teacher & Student & T. & S. & KD & AT & SP & LG & AutoKD & LSKD & \cellcolor{sgray} \textbf{VRM} \vspace{-0.5pt} \\
\midrule
ResNet56 & DeiT-T & 70.43 & 65.08 & 73.25 & 73.51 & 67.36 & 78.15 & \underline{78.58} & 78.55 & \cellcolor{sgray} \textbf{79.52} \vspace{0.5pt} \\
ResNet56 & T2T-ViT-7 & 70.43 & 69.37 & 74.15 & 74.01 & 72.26 & 78.35 & \underline{78.62} & 78.43 & \cellcolor{sgray} \textbf{78.88} \vspace{0.5pt} \\
ResNet56 & PiT-T & 70.43 & 73.58 & 75.47 & 76.03 & 74.97 & 78.48 & 78.51 & \underline{78.76} & \cellcolor{sgray} \textbf{79.25} \vspace{0.5pt} \\
ResNet56 & PVT-T & 70.43 & 69.22 & 74.66 & 77.07 & 70.48 & 77.48 & 73.60 & \underline{78.43} & \cellcolor{sgray} \textbf{79.42} \\
\bottomrule
\end{tabular}} 
\vspace{-1.8mm}
\caption{Results for cross-architecture distillation on CIFAR-100.} \label{tab:cnn_vit} 
\end{minipage} \vspace{-2mm}
\end{table*}

\subsection{Full Objective}
With $\mathcal{G}^{IS}$ and $\mathcal{G}^{IC}$ constructed for both teacher and student predictions, our VRM objective matches edge matrices $\mathcal{E}^{ISV}$ and $\mathcal{E}^{ICV}$ between teacher and student via distance metric $\phi (\cdot)$, for which the Huber loss is used. Formally, $ L^{ISV}_{vrm} = \phi(\mathcal{E}^{ISV}_S, \mathcal{E}^{ISV}_T) $ and $ L^{ICV}_{vrm} = \phi(\mathcal{E}^{ICV}_S, \mathcal{E}^{ICV}_T) $.
The full objective is a weighted combination of the CE loss and the proposed VRM losses:
\begin{equation}
\label{eqn:loss}
 L_{total} = L_{ce}+\alpha L^{ISV}_{vrm} +\beta L^{ICV}_{vrm}
\end{equation}
where $L_{ce}$ is the CE loss applied to student's predictions of both real and virtual views and supervised by GT; $\alpha$ and $\beta$ are balancing scalars. A PyTorch-style pseudo-code for the computing of VRM losses is provided in Algorithm~\ref{alg:vrm} in Supplementary Material.

\section{Experiments}
\label{sec:experiments}

\subsection{Experimental Settings}
Our method is evaluated on CIFAR-100~\cite{cifar100} and ImageNet~\cite{imagenet} datasets for image classification, and MS-COCO~\cite{coco} for object detection. 
All experimental configurations follow the standard practice in prior works. More details are provided in Supplementary Material.

\subsection{Main Results}

\textbf{Results on CIFAR-100.}
The results for different KD methods on CIFAR-100 are shown in Tab.~\ref{tab:cifar100}.
For \textit{same-model KD},
(left of Tab.~\ref{tab:cifar100}), VRM surpasses all previous relation-based methods across all teacher-student pairs by large margins, including DIST~\cite{dist}. 
Noticeably, VRM significantly outperforms the strongest feature-based method, FCFD~\cite{fcfd}.
On average, it also performs much better than top logit-based methods such as CRLD~\cite{crld} and TTM~\cite{ttm}. These are significant, given that IM methods are known to be naturally more adapt at distilling between homogeneous model pairs. 
For \textit{different-model KD} (right of Tab.~\ref{tab:cifar100}), which RM methods are supposed to be more competent with, VRM readily surpasses all RM and IM methods with notable margins.  
These results highlight VRM's versatility on both homogeneous and heterogeneous model pairs.

\vspace{1.6mm}
\noindent \textbf{Results on ImageNet.}
VRM surpasses all feature- and relation-based methods on this large-scale dataset, as shown in Tab.~\ref{tab:imagenet}.
The advantage of VRM is again more apparent over the heterogeneous pair (\textit{i.e.}, ResNet50 $\!\to\!$ MobileNetV2), whereby it for the first time hits 74.0\% Top-1 accuracy. Notably, VRM outperforms strong competitors such as FCFD~\cite{fcfd} and CRLD~\cite{crld} with comparable or even less computational overheads as compared in Tab.~\ref{tab:overhead}.

\begin{figure*}[t] \centering
\begin{minipage}[t]{0.32\textwidth}
\setlength{\tabcolsep}{2pt}
 \centering \scriptsize
\resizebox{0.90\textwidth}{!}{
\begin{tabular}{c|cc}
\toprule
\multirow{2}{*}{Component} & \multirow{2}{*}{\shortstack{RN32$\times$4\\RN8$\times$4}} & \multirow{2}{*}{\shortstack{VGG13\\VGG8}} \\
& & \\
\midrule
Baseline & 72.50 & 70.36 \vspace{-1.3pt} \\
+ Match $\{\mathcal{E}^{IS}\}$ & 75.64 & 74.08 \vspace{-0.5pt} \\
+ Match $\{\mathcal{E}^{IS}, \mathcal{E}^{IC}\}$ & 75.94 & 74.47 \vspace{-0.5pt}  \\
+ Match $\{ \mathcal{E}^{ISV} \}$ & 77.63  & 75.30 \vspace{-0.5pt}  \\
+ Match $\{\mathcal{E}^{ISV},  \mathcal{E}^{ICV}\}$ & 78.38  & 75.56 \vspace{-0.5pt}  \\
\cellcolor{sgray} + Pruning & \cellcolor{sgray} \textbf{78.76} & \cellcolor{sgray} \textbf{76.19} \vspace{-1pt}  \\
\bottomrule
\end{tabular}} 
\vspace{-6pt}
\captionof{table}{Ablation of major designs.}\label{tab:abl_main}
\end{minipage}
\hfill
\begin{minipage}[t]{0.28\textwidth}
\vspace{-13mm}
\centering
\setlength{\tabcolsep}{4.5pt}
\resizebox{\textwidth}{!}{
\begin{tabular}{c|ccc} \toprule
\multirow{2}{*}{\shortstack{Matching\\Choice}} & \multirow{2}{*}{\shortstack{RN32$\times$4\\[1pt]RN8$\times$4}} & \multirow{2}{*}{\shortstack{VGG13\\[1pt]VGG8}} & \multirow{2}{*}{\shortstack{RN50\\[1pt]MNV2}} \\
& & & \\ \midrule
Match $\varnothing$ & 72.50  & 70.36 & 68.33 \vspace{2pt} \\
\cellcolor{sgray} Match $\{\mathcal{E}\}$ & \cellcolor{sgray} \textbf{78.76} & \cellcolor{sgray} \textbf{76.19} & \cellcolor{sgray} \textbf{72.30}  \vspace{2pt} \\
Match $\{\mathcal{E}, \mathcal{V}\}$ & 78.63 & 75.60 & 70.97 \vspace{2pt} \\
Match $\{\mathcal{V}\}$ & 78.46  & 75.77 & 68.33  \\
\bottomrule
\end{tabular}} \vspace{2pt} 
\captionof{table}{Effect of matching vertices.} \label{tab:abl_vm}
\end{minipage} 
\vspace{-5mm}
\hfill
\begin{minipage}[t]{0.36\textwidth}
\vspace{-14mm}
\setlength{\tabcolsep}{0.4pt}
\centering\resizebox{0.92\textwidth}{!}{
\begin{tabular}{ccc|ccc} \toprule
\multirow{2}{*}{$\psi(\cdot)$}  & \multirow{2}{*}{\shortstack{Shape of\\Rel. Matrix}} & \multirow{2}{*}{Method} & \multirow{2}{*}{\shortstack{RN32$\times$4\\RN8$\times$4}} & \multirow{2}{*}{\shortstack{VGG13\\[1pt]VGG8}} & \multirow{2}{*}{\shortstack{RN50\\[1pt]MNV2}} \vspace{-1mm} \\ 
& & & & & \\ \midrule
 \multirow{2}{*}{\shortstack{Gram\\matrices}} & \multirow{2}{*}{\shortstack{$[B,B]$ \& \\ $[C,C]$}} & \multirow{2}{*}{\shortstack{CC, SP, \\ PKT, DIST}} & \multirow{2}{*}{77.31} & \multirow{2}{*}{75.12}  & \multirow{2}{*}{70.28} \vspace{-1mm}  \\ 
 & & & & & \\ \midrule
\multirow{2}{*}{\shortstack{Angular\\relations}} & \multirow{2}{*}{ $[B,B,B]$} & \multirow{2}{*}{RKD} & \multirow{2}{*}{77.14} & \multirow{2}{*}{75.09} & \multirow{2}{*}{69.65} \vspace{-1mm} \\ 
 & & & & & \\ \midrule
\rowcolor{sgray} &  &  &  &  &   \\ 
\rowcolor{sgray}\multirow{-2}{*}{\shortstack{ISV \&\\ ICV}} & \multirow{-2}{*}{\shortstack{$[B,B,C]$ \& \\$[C,C,B]$}} & \multirow{-2}{*}{VRM} & \multirow{-2}{*}{ \textbf{78.46}} & \multirow{-2}{*}{ \textbf{75.30}} & \multirow{-2}{*}{ \textbf{72.10}} \vspace{-1mm} \\ \bottomrule
\end{tabular}} 
\vspace{-1.2mm}
\captionof{table}{Comparing different relation functions.} \label{tab:abl_psi}
\end{minipage} \vspace{5mm}
\end{figure*}

\begin{figure*}[t]
\begin{minipage}[t]{0.31\textwidth}
\centering
\vspace{-3mm}
\setlength{\tabcolsep}{4pt}
\begin{table}[H]
\centering
\resizebox{0.96\linewidth}{!}{
\setlength{\tabcolsep}{6pt}
\begin{tabular}{c|ccccc}
\toprule
$\alpha$ & 32 & 64 & \cellcolor{sgray} 128 & 256 & 512 \vspace{-2pt} \\ \midrule
Acc. & 78.03 & 78.13 & \cellcolor{sgray} \textbf{78.76} & 78.32 & 77.84 \vspace{-2pt} \\
\bottomrule
\end{tabular}}
\end{table}
\vspace{-20pt}
\setlength{\tabcolsep}{2pt}
\begin{table}[H]
\centering
\resizebox{0.96\linewidth}{!}{
\setlength{\tabcolsep}{6pt}
\begin{tabular}{c|ccccc}
\toprule
$\beta$ & 8 & 16 & \cellcolor{sgray} 32 & 64 & 128 \vspace{-2pt} \\ \midrule
Acc. & 78.40 & 78.51 & \cellcolor{sgray} \textbf{78.76} & 78.86 & 78.60 \vspace{-2pt} \\
\bottomrule
\end{tabular}}
\end{table}
\vspace{-20pt}
\begin{table}[H]\centering
\resizebox{0.96\linewidth}{!}{
\setlength{\tabcolsep}{3pt}
\begin{tabular}{c|cccccc}
\toprule
$n$ & 0 & 1 & \cellcolor{sgray} 2 & 3 & 4 & 5 \vspace{-2pt} \\ \midrule
Acc. & 77.92 & 78.38 & \cellcolor{sgray} \textbf{78.76} & 78.55 & 78.43 & 78.50 \vspace{-2pt} \\
\bottomrule
\end{tabular}}
\end{table}
\vspace{-15.5pt}
\captionof{table}{Effect of different $\alpha$, $\beta$, and $n$.} \label{tab:abl_hp} \vspace{-6mm}
\end{minipage}
\hfill
\begin{minipage}[t]{0.31\textwidth}
\centering
\vspace{-0mm}
\includegraphics[width=1.01\linewidth]{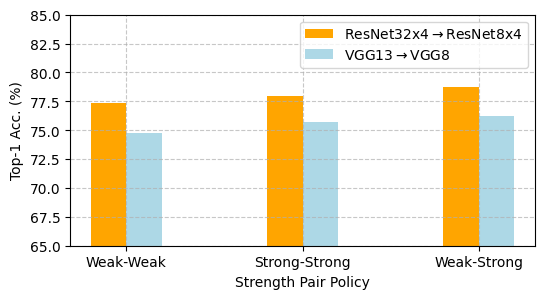} \vspace{-7.8mm}
\caption{Effect of different difficulty pairs.} \label{fig:abl_strength_pair}
\end{minipage}
\hfill
\begin{minipage}[t]{0.36\textwidth}
\centering
\vspace{-0mm}
\includegraphics[width=0.88\linewidth]{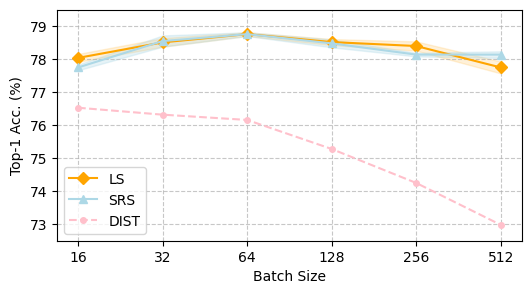} \vspace{-4.1mm}
\caption{Robustness to varying $B$.} \label{fig:abl_b}
\vspace{-15mm}
\end{minipage}
\vspace{-2mm}
\end{figure*}

\begin{figure*}[t]
\centering
\begin{minipage}{0.55\textwidth}
\centering
\subcaptionbox{\scriptsize RKD\label{fig:tsne_rkd}}[0.22\textwidth]{
\includegraphics[width=\linewidth, height=0.94\linewidth]{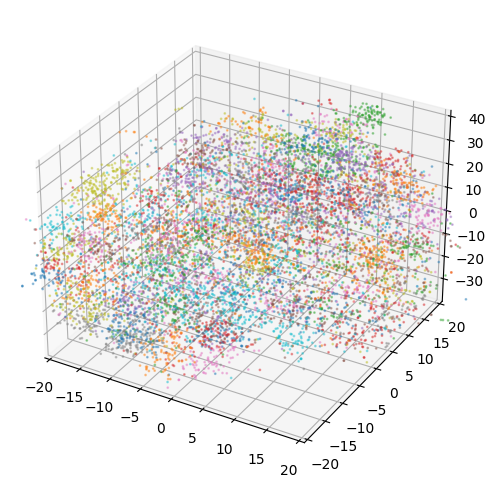}}\hfill
\subcaptionbox{\scriptsize VRM\label{fig:tsne_vrm}}[0.22\textwidth]{
\includegraphics[width=\linewidth, height=0.94\linewidth]{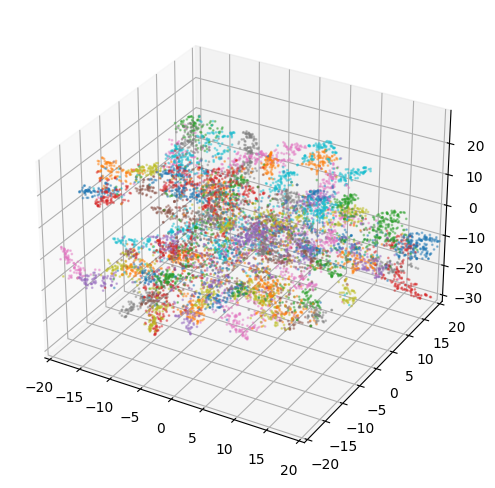}}\hfill
\subcaptionbox{\scriptsize RKD\label{fig:landscape_rkd}}[0.26\textwidth]{\includegraphics[width=0.99\linewidth]{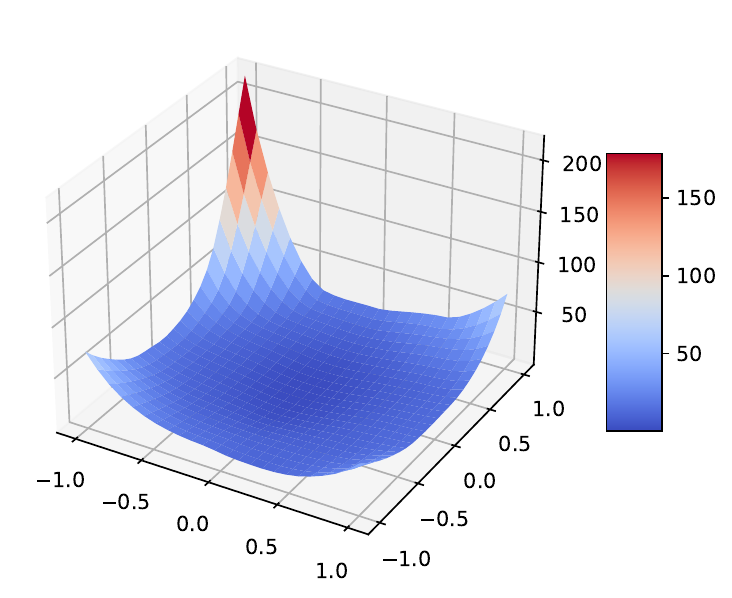}}\hfill
\subcaptionbox{\scriptsize VRM\label{fig:landscape_vrm}}[0.26\textwidth]{
\includegraphics[width=0.99\linewidth]{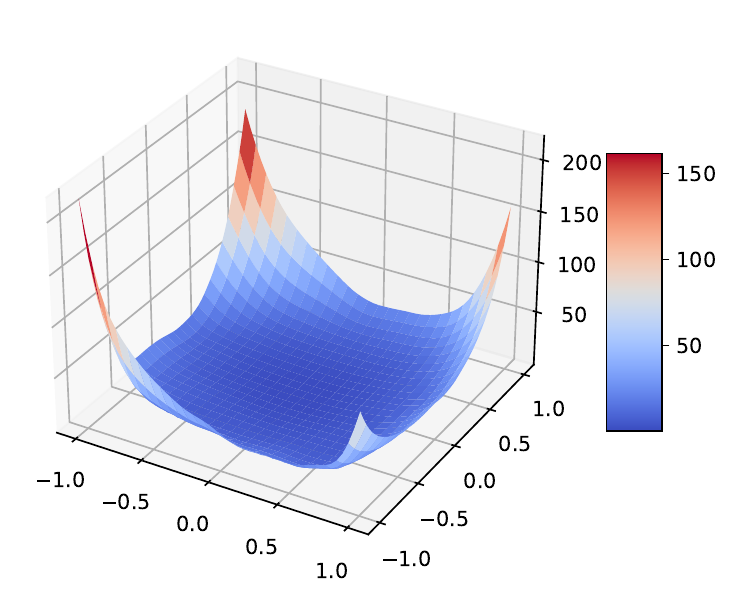}} \vspace{-3mm}
\caption{t-SNE (a–b) and loss landscape (c–d) visualisations for RN8$\times$4 students distilled with an RN32$\times$4 teacher via RKD and VRM.} \label{fig:abl_vis}
\end{minipage}
\hfill
\begin{minipage}[t]{0.41\textwidth}
\centering
\vspace{-14mm}
\resizebox{0.95\textwidth}{!}{
\tabcolsep=0.22cm 
\begin{tabular}{cc|cc}
\toprule
Real & Virtual & RN50 $\!\to\!$ MNV2 & RN34 $\!\to\!$ RN18 \vspace{0.3mm} \\ \midrule
\cellcolor{sgray} Default & \cellcolor{sgray} Default & \cellcolor{sgray} \textbf{74.16}/\textbf{91.78} & \cellcolor{sgray} \textbf{72.67}/90.68  \vspace{0.5mm} \\
Default & RandAug(n=0) & 74.10/91.70 & 72.60/\textbf{90.88}  \vspace{0.5mm} \\
Default & RandAug(n=1) & 74.13/91.63 & 72.59/90.82  \vspace{0.5mm} \\
Default & RandAug(n=2) & 74.13/91.68 & 72.48/90.73  \vspace{0.2mm} \\
\bottomrule
\end{tabular}} \vspace{-1.1mm}
\captionof{table}{Effect of different transformation strength pairs on ImageNet.} \label{tab:aug_in}
\end{minipage}
\label{fig:tsne}
\vspace{-1pt}
\end{figure*}

\begin{figure*}[t]
\begin{minipage}{0.48\textwidth} \centering
\renewcommand{\arraystretch}{0.90}
\resizebox{\textwidth}{!}{
\tabcolsep=0.08cm
\begin{tabular}{c|ccccccccc} \toprule
Method & KD & FitNets & RKD & ICKD & DIST & CRD & ReviewKD \vspace{-1.5pt} \\ \midrule
Train. Time (ms) & 24.9 & 26.7 & 31.0 & 28.0 & 27.2 & 41.2 & 39.9 \vspace{-1pt} \\
Peak GPU Mem. (MB) & 323 & 330 & 330 & 381 & 330 & 1418 & 1042 \vspace{-1pt} \\
\midrule
Method & SDD & MLLD & CRLD & NORM & PEFD & FCFD & \cellcolor{sgray}  \textbf{VRM} \vspace{-1.5pt} \\ \midrule
Train. Time (ms) & 34.2 & 57.2 & 43.2 & 35.1 & 36.2 & 56.4 & \cellcolor{sgray} \textbf{47.2}\vspace{-1pt} \\
\vspace{-2pt}
Peak GPU Mem. (MB) & 690 & 576 & 551 & 1806 & 701 & 953 & \cellcolor{sgray}  \textbf{579} \vspace{-1pt} \\
\bottomrule
\end{tabular}}\vspace{-0.9mm}
\captionof{table}{Training efficiency of different distillation methods.}  \label{tab:overhead} 
\end{minipage}
\hfill
\begin{minipage}{0.48\textwidth} \centering
\renewcommand{\arraystretch}{0.95}
\resizebox{\textwidth}{!}{
\tabcolsep=0.22cm 
\begin{tabular}{c|ccc|cc} \toprule
\multirow{2}{*}{Pruning} & \multicolumn{3}{c|}{ RN32$\times$4 $\!\to\!$ RN8$\times$4} &  {V13 $\!\to\!$ V8} & {W40-2 $\!\to\!$ SNV2} \\ 
\cmidrule(lr){2-4} \cmidrule(lr){5-6}
 & {Acc.} & {$t_{batch}$ (s)} & {Mem. (MB)} & {Acc.} & {Acc.} \\ \midrule
w/o & 78.38 & 61.4 & 34.3 & 75.56 & 78.45 \vspace{0.4mm} \\  
+ REP & 78.46 & \textbf{46.3} & 13.6 & 75.30 & 78.41 \vspace{0.4mm} \\ 
\cellcolor{sgray} + UEP & \cellcolor{sgray} \textbf{78.76} & \cellcolor{sgray} 47.2 & \cellcolor{sgray} \textbf{13.1} & \cellcolor{sgray} \textbf{76.19} & \cellcolor{sgray} \textbf{78.62} \vspace{-1pt} \\ 
\bottomrule 
\end{tabular}} 
 \vspace{-0.6mm}
\captionof{table}{Effectiveness and efficiency of pruning operations.} \label{tab:pruning} 
\end{minipage}
\vspace{-2mm}
\end{figure*}

\vspace{1.6mm}
\noindent \textbf{Results on MS-COCO.}
We demonstrate that VRM generalises to more challenging tasks by adapting it to object detection. From Tab.~\ref{tab:coco}, VRM performs better than existing logit-based methods and competitively to feature-based methods. VRM is slightly behind top-performing feature-based methods such as FCFD~\cite{fcfd} and ReviewKD~\cite{reviewkd}. This is a consequence of the very nature of the object detection task, where fine-grained contextual features play a vital role, making feature-based methods inherently better-off~\cite{fgfi}. Nonetheless, we experimentally demonstrate that VRM is effective on object detection, whereby it improves the baseline by $2\text{--}3\%$ AP, surpasses all logit-based methods, and is on par with strong feature-based methods.

\vspace{1.6mm}
\noindent \textbf{Results for cross-architecture distillation.}
We follow the settings in~\cite{tinytransformer} and~\cite{lskd} to perform CNN-to-ViT cross-architecture distillation.
Tab.~\ref{tab:cnn_vit} shows that VRM sets new record under this set-up for different ViT architectures. Notably, by simply replacing the vanilla KD loss with the proposed VRM objective and making no other modifications, VRM improves DeiT-T's accuracy by \textit{14.44\%}, leading the relation-based SP~\cite{spkd} by \textit{12.16\%} and even surpassing AutoKD~\cite{autokd} with AutoML search. 

\subsection{Ablation Studies} \label{sec:ablations}

\noindent \textbf{Ablations of main design choices.}
In Tab.~\ref{tab:abl_main}, starting from the baseline where only the CE loss is applied, we gradually incorporate each of our designs. As shown, each extra design consistently brings performance gains, which corroborate the validity of individual design choices.

\vspace{1.6mm}
\noindent \textbf{Effect of different relation encoding functions.}
Tab.~\ref{tab:abl_psi} compares our proposed relation encoding scheme with existing ones. To demonstrate the superiority of our formulation, we substitute it with existing Gram matrices~\cite{cckd, spkd, pkt, dist} or angle-wise relations~\cite{rkd}. Results show that our inter-sample class-wise and inter-class sample-wise relations consistently yield much better performance than existing relations, which highlights VRM's contribution of a novel and superior type of relation encoding in addition to the introduction of virtual relation.

\vspace{1.6mm}
\noindent \textbf{Effect of varying hyperparameters.} 
VRM involves three major hyperparameters: $\alpha$ and $\beta$, and $n$. According to Tab.~\ref{tab:abl_hp}, VRM works generally well with $\alpha$ and $\beta$ in a reasonable range. Larger $\alpha$ and $\beta$ may produce better results for certain model pairs but worse results for others. For $n$, a larger $n$ means more difficult images and larger inter-view prediction discrepancy. From Tab.~\ref{tab:abl_hp}, $n=2$ works best, whereas other values also report competitive results.

\vspace{1.6mm}
\noindent \textbf{Effect of different real-virtual difficulty pairs.}
We also experiment with different strength combinations for generating the real and virtual samples. The results are presented in Fig.~\ref{fig:abl_strength_pair}, where ``Weak'' denotes the default transformation used in previous methods, and ``Strong'' denotes RandAugment with $n=2$.
Overall, we conclude that 1) a moderate discrepancy leads to optimal performance, and 2) the discrepancy is significant to the success of VRM.

\vspace{1.6mm}
\noindent \textbf{Robustness to varying batch sizes.} Relational methods are known to be sensitive to training batch size $B$. We conduct experiments to examine how robust VRM is against varying $B$. 
To adjust the learning rate accordingly, we consider two LR scaling rules: linear LR scaling~\cite{imnet1h}  and square root LR scaling~\cite{simclr}.
From Fig.~\ref{fig:abl_b}, VRM yields competitive results across a wide range of $B$ values. 
In contrast, DIST~\cite{dist} degrades significantly as $B$ grows.

\subsection{Further Analysis}

\noindent \textbf{Analysis of training dynamics.}
Figs.~\ref{fig:train_dynamics} and ~\ref{fig:val_dynamics} plot the per-epoch training and validation set accuracies throughout training. VRM maintains a all-time lead in both training and validation performance, with faster convergence. Besides, while VRM's lead in training performance tapers off towards the end of training, it remains more substantial, if not further enlarging, in validation performance, which highlights its superior generalisation properties. This analysis also shows our designs are effective in mitigating the issues with existing RM methods identified in the pilot studies. 

\vspace{1.6mm}
\noindent  \textbf{Analysis of spurious gradients.} We revisit our second pilot study by analysing how spurious gradients are suppressed in real ImageNet training. From Fig.~\ref{fig:pilot_in}, similar gradient diffusion patterns are observed in real distillation. Samples semantically more similar to the sample with spurious predictions are more affected by spurious gradients (\textit{e.g.}, the dog image with a watery background resembles the bubbly texture in another sample that is actually a computer mouse.). 

\vspace{1.6mm}
\noindent \textbf{Visualisation of embedding space.} We conduct t-SNE analysis on student penultimate layer embeddings learnt via different methods. As presented in Fig.~\ref{fig:tsne_rkd} and~\ref{fig:tsne_vrm}, VRM leads to more compact per-class clusters with clearer inter-class separation and less stray points. These imply better student features distilled by VRM for the downstream task. 

\vspace{1.6mm}
\noindent  \textbf{Visualisation of loss landscape.} We further analyse the generalisation and convergence properties of our method through the lens of visualised loss landscape~\cite{losslandscape}.  
From Fig.~\ref{fig:landscape_rkd} and~\ref{fig:landscape_vrm}, compared to RKD, VRM has a wider, flatter, and deeper region of minima -- typical hint of better model generalisation and robustness; this wide convexity basin surrounded by salient pikes in various directions indicates excellent convergence properties. 
More visualisations are provided in Supplementary Material.

\vspace{1.6mm}
\noindent \textbf{Vertex matching.} 
Within the constructed graphs $\mathcal{G}^{IS}$ and $\mathcal{G}^{IC}$, 
VRM matches edges $\mathcal{E}^{ISV}$ and $\mathcal{E}^{ICV}$ which carry relational knowledge. By extension, we can naturally expect vertices $\mathcal{V}$ to be also transferred. 
From Tab.~\ref{tab:abl_vm}, matching vertices is not as effective. The advantage of matching relations (\textit{i.e.}, edges) is more pronounced for heterogeneous KD pairs such as ResNet50 $\!\to\!$ MobileNetV2. Adding vertex matching to the proposed edge matching does not improve but instead degrade performance. We argue this is because introducing vertex matching objectives make the matching criterion more stringent which is against our motivation of using more slackened matching.

\vspace{1.6mm}
\noindent \textbf{Role of transformation operations.} We emphasise the strong performance of VRM is not a simple outcome of the transformations involved. This can be verified by the results in Fig.~\ref{fig:abl_strength_pair} and Tab.~\ref{tab:aug_in}, where using only the default weak transformation (\textit{i.e.}, random crop and horizontal flip) for both views achieves highly competitive and even superior performance under VRM. In fact, in the case of ``Weak-Weak'', both views still differ because of the stochastic operations used. 
Essentially, \textit{the success of VRM lies exactly in the discrepancy of teacher and student predictions}, where regularisation comes into crucial play. On smaller and easier datasets such as CIFAR-100, the gap between teacher and student predictions are relatively small, and manually enlarging this gap via extra  transformations leads to promising outcomes. Whereas the more difficult ImageNet data already sees large teacher-student prediction discrepancy, such that it benefits less from further amplification.

\vspace{1.6mm}
\noindent  \textbf{Training efficiency.} From Tab.~\ref{tab:overhead}, VRM is reasonably efficient compared to existing algorithms, including some leading feature-~\cite{norm, pefd, fcfd} and logit-based~\cite{sdd, mlld} methods.
VRM does not introduce any extra overheads at inference.
We also breakdown the proposed pruning designs in Tab.~\ref{tab:pruning} to understand how they individually impact the performance and efficiency of VRM. We see REP boosts efficiency at minimal performance cost, whereas UEP trades minimal efficiency cost for performance. 

\vspace{1.6mm}
\noindent \textbf{More experiments, analyses, and discussions} are provided in Supplementary Material.

\section{Conclusion}
This paper presents VRM, a novel knowledge distillation framework that constructs and transfers virtual relations. Our designs are motivated by a set of pilot experiments, from which we identified two main cruxes with existing relation-based KD methods: their tendency to overfit and susceptibility to adverse gradient propagation. A series of tailored designs are developed and are shown to successfully mitigate these issues.
We have conducted extensive experiments on different tasks and multiple datasets and verified VRM's validity and superiority in diverse settings, whereby it consistently achieves state-of-the-art performance. We hope that this work could renew the community's interest in relation-based knowledge distillation, and encourage more systematic reassessment of the design principles of such solutions.

\newpage
\paragraph{Acknowledgement.} This work was supported in part by NSFC (62322113, 62376156), Shanghai Municipal Science and Technology Major Project (2021SHZDZX0102), and the Fundamental Research Funds for the Central Universities.

{
    \small
    \bibliographystyle{ieeenat_fullname}
    \bibliography{main}
}

\clearpage
\setcounter{page}{1}
\maketitlesupplementary

\section{Appendix}
\subsection{List of All Compared Methods}
A list of all methods we have compared with in this paper is as follows:

\textbf{Feature-based} methods include FitNets~\cite{fitnets}, NST~\cite{nst}, AT~\cite{at}, AB~\cite{ab}, OFD~\cite{ofd}, VID~\cite{vid}, CRD~\cite{crd}, SRRL~\cite{srrl}, SemCKD~\cite{semckd}, PEFD~\cite{pefd}, MGD~\cite{mgd}, CAT-KD~\cite{catkd}, TaT~\cite{tat}, ReviewKD~\cite{reviewkd}, NORM~\cite{norm}, FCFD~\cite{fcfd}, and RSD~\cite{rsd}. \vspace{0.5mm}

\textbf{Logit-based} methods include KD~\cite{kd}, DML~\cite{dml}, TAKD~\cite{takd}, CTKD~\cite{ctkd}, NKD~\cite{nkd}, DKD~\cite{dkd}, LSKD~\cite{lskd}, TTM~\cite{ttm}, SDD~\cite{sdd}, and CRLD~\cite{crld}. \vspace{0.5mm}

\textbf{Relation-based} methods include FSP~\cite{fsp}, RKD~\cite{rkd}, PKT~\cite{pkt}, CC~\cite{cckd}, SP~\cite{spkd}, ICKD~\cite{ickd}, and DIST~\cite{dist}. 

For MS-COCO object detection, we also compare VRM with FGFI~\cite{fgfi}. For ConvNet-to-ViT experiments, we also present the results for LG~\cite{tinytransformer} and AutoKD~\cite{autokd}.

\subsection{List of All Transformation Operations} \label{sec:appendix_transforms}
For our main experiments, we borrow the RandAugment implementation from the \texttt{TorchSSL} codebase\footnote{\url{https://github.com/TorchSSL}}.
It comprises a total of 14 image transformation operations, namely: \vspace{0.5mm}
\begin{enumerate}
    \item \texttt{Autocontrast}: automatically adjust image contrast \vspace{0.5mm}
    \item \texttt{Brightness}: adjust image brightness \vspace{0.5mm}
    \item \texttt{Color}: adjust image colour balance \vspace{0.5mm}
    \item \texttt{Contrast}: adjust image contrast \vspace{0.5mm}
    \item \texttt{Equalize}: equalise image histogram \vspace{0.5mm}
    \item \texttt{Identity}: leave image unaltered \vspace{0.5mm}
    \item \texttt{Posterize}: reduce number of bits for each channel \vspace{0.5mm}
    \item \texttt{Rotate}: rotate image \vspace{0.5mm}
    \item \texttt{Sharpness}: adjust image sharpness \vspace{0.5mm}
    \item \texttt{Shear\_x}: shear image horizontally \vspace{0.5mm}
    \item \texttt{Shear\_y}: shear image vertically \vspace{0.5mm}
    \item \texttt{Solarize}: invert all pixels above a threshold \vspace{0.5mm}
    \item \texttt{Translate\_x}: translate image horizontally \vspace{0.5mm}
    \item \texttt{Translate\_y}: translate image vertically \vspace{1mm}
\end{enumerate}

Besides, we also apply \texttt{Cutout} with a probability of $1.0$, which sets a square patch of random size within the image to gray. The above operations are preceded by \texttt{RandomCrop} and \texttt{RandomHorizontalFlip} in our virtual view image generation pipeline.

For ConvNet-to-ViT experiments, we follow~\citet{tinytransformer} and use the RandAugment function provided by the \texttt{timm} library~\footnote{\url{https://github.com/huggingface/pytorch-image-models}}. This function contains 15 image transformation operations:
\begin{enumerate} \vspace{0.5mm}
    \item \texttt{AutoContrast}: automatically adjust image contrast \vspace{0.5mm}
    \item \texttt{Brightness}: adjust image brightness \vspace{0.5mm}
    \item \texttt{Color}: adjust image colour balance \vspace{0.5mm}
    \item \texttt{Contrast}: adjust image contrast \vspace{0.5mm}
    \item \texttt{Equalize}: equalise image histogram \vspace{0.5mm}
    \item \texttt{Invert}: invert image \vspace{0.5mm}
    \item \texttt{Posterize}: reduce number of bits for each channel \vspace{0.5mm}
    \item \texttt{Rotate}: rotate image \vspace{0.5mm}
    \item \texttt{Sharpness}: adjust image sharpness \vspace{0.5mm}
    \item \texttt{ShearX}: shear image horizontally \vspace{0.5mm}
    \item \texttt{ShearY}: shear image vertically \vspace{0.5mm}
    \item \texttt{Solarize}: invert all pixels above a threshold \vspace{0.5mm}
    \item \texttt{SolarizeAdd}: add a certain value to all pixels below a threshold \vspace{0.5mm}
    \item \texttt{TranslateXRel}: translate image horizontally by a fraction of its width \vspace{0.5mm}
    \item \texttt{TranslateYRel}: translate image vertically by a fraction of its height \vspace{1mm}
\end{enumerate}

\let\oldtexttt\texttt
\renewcommand{\texttt}{\normalfont} 
\begin{algorithm}[t]
\caption{PyTorch-style pseudo-code for computing VRM loss} \label{alg:vrm}
\begin{algorithmic}[1] \scriptsize
\STATE \textcolor{codegreen}{\# x: a batch of raw samples}
\STATE \textcolor{codegreen}{\# T$_r$, T$_v$: transformation functions for real and virtual views}
\STATE \textcolor{codegreen}{\# f$^T$, f$^S$: teacher and student networks}
\STATE \textcolor{codegreen}{\# B: batch size, C: number of classes} 

\vspace{2mm}
\STATE \textcolor{codegreen}{\# Generate real and virtual views of samples}
\STATE x$_r$, x$_v$ = T$_r$(x), T$_v$(x) \textcolor{codegreen}{\# [B]}

\vspace{2mm}
\STATE \textcolor{codegreen}{\# Obtain teacher and student predictions}
\STATE z$^T_r$, z$^T_v$ = f$^T$(x$_r$), f$^T$(x$_v$) \textcolor{codegreen}{\# [B,C]}

\STATE z$^S_r$, z$^S_v$ = f$^S$(x$_r$), f$^S$(x$_v$) \textcolor{codegreen}{\# [B,C]}
\vspace{2mm}
\STATE \textcolor{codegreen}{\# Compute inter-sample virtual edge matrices}
\STATE e$^T_{\text{ISV}}$ = Norm(z$^T_r$.unsqueeze(0) - z$^T_v$.unsqueeze(1)) \textcolor{codegreen}{\# [B,B,C]}
\STATE e$^S_{\text{ISV}}$ = Norm(z$^S_r$.unsqueeze(0) - z$^S_v$.unsqueeze(1)) \textcolor{codegreen}{\# [B,B,C]}

\vspace{2mm}
\STATE \textcolor{codegreen}{\# Compute inter-class virtual edge matrices}
\STATE e$^T_{\text{ICV}}$ = Norm(z$^T_r$.unsqueeze(1) - z$^T_v$.unsqueeze(2)) \textcolor{codegreen}{\# [C,C,B]}
\STATE e$^S_{\text{ICV}}$ = Norm(z$^S_r$.unsqueeze(1) - z$^S_v$.unsqueeze(2)) \textcolor{codegreen}{\# [C,C,B]}

\vspace{2mm}
\STATE \textcolor{codegreen}{\# Obtain unreliable edge masks}
\STATE M$_{\text{ISV}}$ = JE(z$^S_r$.unsqueeze(0), z$^S_v$.unsqueeze(1)) $<$ p$_m$ \textcolor{codegreen}{\# [B,B]}
\STATE M$_{\text{ICV}}$ = JE(z$^S_r$.unsqueeze(1), z$^S_v$.unsqueeze(2)) $<$ p$_m$ \textcolor{codegreen}{\# [C,C]}

\vspace{1.5mm}
\STATE \textcolor{codegreen}{\# Compute VRM losses}
\STATE loss\_isv = (MSE(e$^T_{\text{ISV}}$, e$^S_{\text{ISV}}$) * M$_{\text{ISV}}$).sum()
\STATE loss\_icv = (MSE(e$^T_{\text{ICV}}$, e$^S_{\text{ICV}}$) * M$_{\text{ICV}}$).sum() 

\vspace{2mm}
\STATE \textcolor{codeblue}{return} loss\_isv, loss\_icv
\end{algorithmic}
\end{algorithm}
\let\texttt\oldtexttt

\begin{table*}[t]
\setlength{\tabcolsep}{3pt}
\centering\resizebox{0.68\textwidth}{!}{
\begin{tabular}{cccccccc}
\toprule
\multirow{2}{*}{\shortstack{Teacher \\ [2pt]Student}} & \multirow{4}{*}{Venue} & \multirow{2}{*}{\shortstack{ResNet56 \\ [2pt]ResNet20}} & \multirow{2}{*}{\shortstack{ResNet110 \\ [2pt]ResNet32}} & \multirow{2}{*}{\shortstack{ResNet32$\times$4 \\[2pt]ResNet8$\times$4}} & \multirow{2}{*}{\shortstack{WRN-40-2 \\ [2pt]WRN-16-2}} & \multirow{2}{*}{\shortstack{WRN-40-2 \\ [2pt]WRN-40-1}} & \multirow{2}{*}{\shortstack{VGG13 \\ [2pt]VGG8}} \\
& & & & & & & \\
\cmidrule(lr){1-1} \cmidrule(lr){3-8}
Teacher & & 72.34 & 74.31 & 79.42 & 75.61 & 75.61 & 74.64 \\
Student & & 69.06 & 71.14 & 72.50 & 73.26 & 71.98 & 70.36 \\
\midrule
\multicolumn{8}{c}{\textit{Feature-based}}\vspace{1mm} \\
FitNets~\cite{fitnets} & ICLR'15 & 69.21 & 71.06 & 73.50 & 73.58 & 72.24 & 71.02 \\
NST~\cite{nst} & arXiv'17 & 69.60 & 71.96 & 73.30 & 73.68 & 72.24 & 71.53 \\
AT~\cite{at} & ICLR'17 & 70.55 & 72.31 & 73.44 & 74.08 & 72.77 & 71.43 \\
AB~\cite{ab} & AAAI'19 & 69.47 & 70.98 & 73.17 & 72.50 & 72.38 & 70.94 \\
OFD~\cite{ofd} & ICCV'19 & 70.98 & 73.23 & 74.95 & 75.24 & 74.33 & 73.95 \\
VID~\cite{vid} & CVPR'19 & 70.38 & 72.61 & 73.09 & 74.11 & 73.30 & 71.23 \\
CRD~\cite{crd} & ICLR'20 & 71.16 & 73.48 & 75.51 & 75.48 & 74.14 & 73.94 \\
SRRL~\cite{srrl} & ICLR'21 & 71.13 & 73.48 & 75.33 & 75.59 & 74.18 & 73.44 \\
PEFD~\cite{pefd} & NeurIPS'22 & 70.07 & 73.26 & 76.08 & 76.02 & 74.92 & 74.35 \\
CAT-KD~\cite{catkd} & CVPR'23 & 71.05 & 73.62 & 76.91 & 75.60 & 74.82 & 74.65 \\
TaT~\cite{tat} & CVPR'22 & 71.59 & 74.05 & 75.89 & 76.06 & 74.97 & 74.39 \\
ReviewKD~\cite{reviewkd} & CVPR'21 & 71.89 & 73.89 & 75.63 & 76.12 & 75.09 & 74.84 \\
NORM~\cite{norm} & ICLR'23 & 71.35 & 73.67 & 76.49 & 75.65 & 74.82 & 73.95 \\
FCFD~\cite{fcfd} & ICLR'23 & 71.96 & - & 76.62 & 76.43 & 75.46 & 75.22 \\

\midrule
\multicolumn{8}{c}{\textit{Logit-based}}\vspace{1mm} \\
KD~\cite{kd} & arXiv'15 & 70.66 & 73.08 & 73.33 & 74.92 & 73.54 & 72.98  \\
DML~\cite{dml} & CVPR'18 & 69.52 & 72.03 & 72.12 & 73.58 & 72.68 & 71.79 \\
TAKD~\cite{takd} & AAAI'20 & 70.83 & 73.37 & 73.81 & 75.12 & 73.78 & 73.23 \\
CTKD~\cite{ctkd} & AAAI'23 & 71.19 & 73.52 & 73.79 & 75.45 & 73.93 & 73.52 \\
NKD~\cite{nkd} & ICCV'23 & 70.40 & 72.77 & 76.35 & 75.24 & 74.07 & 74.86 \\
DKD~\cite{dkd} & CVPR'22 & 71.97 & 74.11 & 76.32 & 76.24 & 74.81 & 74.68 \\
LSKD~\cite{lskd} & CVPR'24 & 71.43 & 74.17 & 76.62 & 76.11 & 74.37 & 74.36 \\
TTM~\cite{ttm} & ICLR'24 & 71.83 & 73.97 & 76.17 & 76.23 & 74.32 & 74.33 \\
CRLD~\cite{crld} & MM'24 & 72.10 & 74.42 & 77.60 & 76.45 & 75.58 & 75.27 \\
\midrule
\multicolumn{8}{c}{\textit{Relation-based}}\vspace{1mm} \\
FSP~\cite{fsp} & CVPR'17 & 69.95 & 71.89 & 72.62 & 72.91 & - & 70.20 \\
RKD~\cite{rkd} & CVPR'19 & 69.61 & 71.82 & 71.90 & 73.35 & 72.22 & 71.48 \\
PKT~\cite{pkt} & ECCV'18 & 70.34 & 72.61 & 73.64 & 74.54 & 73.45 & 72.88 \\
CCKD~\cite{cckd} & CVPR'19 & 69.63 & 71.48 & 72.97 & 73.56 & 72.21 & 70.71  \\
SP~\cite{spkd} & ICCV'19 & 69.67 & 72.69 & 72.94 & 73.83 & 72.43 & 72.68  \\
ICKD~\cite{ickd} & ICCV'21 & 71.76 & 73.89 & 75.25 & 75.64 & 74.33 & 73.42 \\
DIST~\cite{dist} & NeurIPS'22 & 71.75 & - & 76.31 & - & 74.73 & - \\
\cellcolor{sgray} \textbf{VRM} & \cellcolor{sgray} - & \cellcolor{sgray} \textbf{72.09} & \cellcolor{sgray} \textbf{75.03} & \cellcolor{sgray} \textbf{78.76} & \cellcolor{sgray} \textbf{77.47} & \cellcolor{sgray} \textbf{76.46} & \cellcolor{sgray} \textbf{76.19} \\
\bottomrule
\end{tabular}}
\caption{Top-1 accuracy (\%) on CIFAR-100 for same-model teacher-student pairs.}
\label{tab:cifar100_same}
\end{table*}

Similar to the role of \texttt{Cutout}, the \texttt{timm} library additionally implements a \texttt{RandomErasing} operation, which sets a  rectangular patch of random size and shape within the image to random pixels. The above operations are preceded by \texttt{RandomResizedCropAndInterpolation} and \texttt{RandomHorizontalFlip} in our strong view image generatino pipeline, which is the default configuration in \texttt{timm}.

\subsection{Pseudo-Code}
In Algorithm~\ref{alg:vrm}, we provide the PyTorch-style pseudo-code for the calculation of the proposed VRM losses. The construction of relation edges can be conveniently implemented through some matrix operations, with the redundancy edge pruning implicitly incorporated. Overall, the proposed losses can be neatly implemented in about 10 lines of codes in PyTorch. 

\subsection{Details on Experimental Configurations} \label{sec:appendix_implem}

\textbf{Datasets.} 
We conduct experiments on CIFAR-100 and ImageNet for image classification, and MS-COCO for object detection.
CIFAR-100~\cite{cifar100} contains 60k $32\times32$ RGB images annotated in 100 classes. It is split into 50,000 training and 10,000 validation images. ImageNet~\cite{imagenet} is a 1,000-category large-scale image recognition dataset. It provides 1.28 million RGB images for training and 5k for validation. MS-COCO~\cite{coco} is an object detection dataset with images of  common objects in 80 categories. We experiment with its \texttt{train2017} and \texttt{val2017} that include 118k training and 5k validation images, respectively.

\begin{table*}[t]
\setlength{\tabcolsep}{3pt}
\centering\resizebox{0.78\textwidth}{!}{
\begin{tabular}{cccccccc}
\toprule
\multirow{2}{*}{\shortstack{Teacher \\ [2pt]Student}}  & \multirow{4}{*}{Venue} & \multirow{2}{*}{\shortstack{ResNet32$\times$4 \\ [2pt]ShuffleNetV2}} & \multirow{2}{*}{\shortstack{VGG13 \\ [2pt]MobileNetV2}} &  \multirow{2}{*}{\shortstack{ResNet50 \\ [2pt]MobileNetV2}} & \multirow{2}{*}{\shortstack{ResNet50 \\ [2pt]VGG8}} & \multirow{2}{*}{\shortstack{ResNet32$\times$4 \\ [2pt]ShuffleNetV1}} & \multirow{2}{*}{\shortstack{WRN-40-2 \\ [2pt]ShuffleNetV1}} \\
& & & & & & & \\
\cmidrule(lr){1-1} \cmidrule(lr){3-8}
Teacher & & 79.42 & 74.64 & 79.34 & 79.34 & 79.42 & 75.61 \\
Student & & 71.82 & 64.60 & 64.60 & 70.36 & 70.50 & 70.50 \\
\midrule
\multicolumn{8}{c}{\textit{Feature-based}}\vspace{1mm} \\
FitNets~\cite{fitnets} & ICLR'15 & 73.54 & 64.16 & 63.16 & 70.69 & 73.59 & 73.73 \\
NST~\cite{nst} & arXiv'17 & 74.68 & 58.16 & 64.96 & 71.28 & 74.12 & 74.89 \\
AB~\cite{ab} & AAAI'19 & 74.31 & 66.06 & 67.20 & 70.65 & 73.55 & 73.34 \\
AT~\cite{at} & ICLR'17 & 72.73 & 59.40 & 58.58 & 71.84 & 71.73 & 73.32 \\
VID~\cite{vid} & CVPR'19 & 73.40 & 65.56 & 67.57 & 70.30 & 73.38 & 73.61 \\
OFD~\cite{ofd} & ICCV'19 & 76.82 & 69.48 & 69.04 & - & 75.98 & 75.85 \\
CRD~\cite{crd} & ICLR'20 & 75.65 & 69.63 & 69.11 & 74.30 & 75.11 & 76.05 \\
MGD~\cite{mgd}& ECCV'22 & 76.65 & 69.44 & 68.54 & 73.89 & 76.22 & 75.89 \\
SemCKD~\cite{semckd} & AAAI'21 & 77.02 & 69.98 & 68.69 & 74.18 & 76.31 & 76.06 \\
ReviewKD~\cite{reviewkd}& CVPR'21 & 77.78 & 70.37 & 69.89 & 75.34 & 77.45 & 77.14 \\
NORM~\cite{norm} & ICLR'23 & 78.32 & 69.38 & 71.17 & 75.67 & 77.79 & 77.63 \\
FCFD~\cite{fcfd} & ICLR'23 & 78.18 & 70.65 & 71.00 & - & 78.12 & 77.99 \\
CAT-KD~\cite{catkd} & CVPR'23 & 78.41 & 69.13 & 71.36 & - & 78.26 & 77.35 \\
\midrule
\multicolumn{8}{c}{\textit{Logit-based}}\vspace{1mm} \\
KD~\cite{kd} & arXiv'15 & 74.45 & 67.37 & 67.35 & 73.81 & 74.07 & 74.83 \\
DML~\cite{dml} & CVPR'18 & 73.45 & 65.63 & 65.71 & - & 72.89 & 72.76 \\
TAKD~\cite{takd} & AAAI'20 & 74.82 & 67.91 & 68.02 & - & 74.53 & 75.34 \\
CTKD~\cite{ctkd} & AAAI'23 & 75.31 & 68.46 & 68.47 & - & 74.48 & 75.78 \\
NKD~\cite{nkd} & ICCV'23 & 76.26 & 70.22 & 70.76 & 74.01 & 75.31 & 75.96 \\
DKD~\cite{dkd} & CVPR'22 & 77.07 & 69.71 & 70.35 & - & 76.45 & 76.70 \\
LSKD~\cite{lskd} & CVPR'24 & 75.56 & 68.61 & 69.02 & - & - & - \\
TTM~\cite{ttm} & ICLR'24 & 76.55 & 69.16 & 69.59 & 74.82 & 74.37 & 75.42 \\
SDD~\cite{sdd} & CVPR'24 & 76.67 & 68.79 & 69.55 & 74.89 & 76.30 & 76.54 \\ 
CRLD~\cite{crld} & MM'24 & 78.27 & 70.39 & 71.36 & - & - & - \\ 
\midrule
\multicolumn{8}{c}{\textit{Relation-based}}\vspace{1mm} \\
RKD~\cite{rkd} & CVPR'19 & 73.21 & 64.52 & 64.43 & 71.50 & 72.28 & 72.21 \\
PKT~\cite{pkt} & ECCV'18 & 74.69 & 67.13 & 66.52 & 73.01 & 74.10 & 73.89 \\
CCKD~\cite{cckd} & CVPR'19 & 71.29 & 64.86 & 65.43 & 70.25 & 71.14 & 71.38 \\
SP~\cite{spkd} & ICCV'19 & 74.56 & 66.30 & 68.08 & 73.34 & 73.48 & 74.52 \\
DIST~\cite{dist} & NeurIPS'22 & 77.35 & 68.50 & 68.66 & 74.11 & 76.34 & 76.40 \\
\cellcolor{sgray} \textbf{VRM} & \cellcolor{sgray} - & \cellcolor{sgray} \textbf{79.34} & \cellcolor{sgray} \textbf{71.66} & \cellcolor{sgray}  \textbf{72.30} & \cellcolor{sgray}  \textbf{76.96} & \cellcolor{sgray}  \textbf{78.28} & \cellcolor{sgray} \textbf{78.62} \\
\bottomrule
\end{tabular}}
\caption{Top-1 accuracy (\%) on CIFAR-100 for different-model teacher-student pairs.}
\label{tab:cifar100_diff}
\end{table*}

\paragraph{Configurations for main experiments.} 
For CIFAR-100 main experiments, we strictly follow the standard training configurations in previous works~\cite{fcfd, dkd, lskd}. Specifically, we train our framework for 240 epochs using the SGD optimiser and a batch size of 64. The initial LR is 0.01 for MobileNets~\cite{mobilenet} and ShuffleNets~\cite{shufflenet} and 0.05 for other architectures, which decay by a factor of 10 at [150th, 180th, 210th] epochs. Momentum and weight decay are set to 0.9 and 5e-4, respectively. Softmax temperature is set to 4.

For ImageNet experiments, as per standard practice, we train our framework for 100 epochs with a batch size of 256 on two GPUs, with an initial LR of 0.1 that decays by a factor of 10 at [30th, 60th, 90th] epochs. Momentum and weight decay are set to 0.9 and 1e-4, respectively. Softmax temperature is set to 2.

For MS-COCO object detection, we adopt the configurations of~\citet{fgfi, reviewkd, dkd, lskd, norm} whereby we experiment with Faster-RCNN-FPN~\cite{fpn} with different backbone models.
All models are trained for 180,000 iterations on 2 GPUs with a batch size of 8. The LR is initially set as 0.01 and decays at the 120,000th and 160,000th iterations.

\paragraph{Configurations for ConvNet-to-ViT experiments.}
As few studies have considered this setting, we developed our experiments following ~\cite{lskd, autokd} on the codebase provided by ~\cite{tinytransformer}. Our experimental configurations follow~\cite{tinytransformer} and ~\cite{drloc}. Specifically, the ResNet56 teacher is trained for 300 epochs with an initial LR of 0.1 and a cosine LR schedule. The resulted pretrained teacher has a top-1 accuracy of 71.61\%.
All ViTs are trained for 300 epochs (including 20-epoch linear warm-up) using the AdamW optimiser. The initial LR is 5e-4 with a weight decay of 0.05, which eventually decays to 5e-6 via a cosine LR policy. The ResNet56 teacher is trained on $32\times32$ resolution images, while ViT students are fed with $224\times224$ images. The default RandAugment is applied for data augmentation, with number of randomly sampled operations $n$ set to 2, transform magnitude $m$ to 9, and probability of applying random erasing $p$ to 0.25. All models are trained on a single NVIDIA RTX 3090 GPU with a batch size of 128.

\paragraph{Implementations.} 
Our method is implemented in the \texttt{mdistiller}\footnote{\url{https://github.com/megvii-research/mdistiller}} codebase in PyTorch for image classification experiments. For object detection, it also partially builds upon the \texttt{detectron2}\footnote{\url{https://github.com/facebookresearch/detectron2}} library. For ConvNet-to-ViT experiments, we utilise the  \texttt{pycls}\footnote{\url{https://github.com/facebookresearch/pycls}} and the \texttt{tiny-transformer}\footnote{\url{https://github.com/lkhl/tiny-transformers}} codebases. All reported results are average over 3 trials.

\paragraph{Efficiency benchmarking.}
Tab.~\ref{tab:overhead} benchmarks the training time per batch and the peak GPU memory usage of various methods on a workstation equipped with 20 Intel Core i9-10850K CPUs (10 cores) and an NVIDIA RTX 3090 GPU. All measurements are taken on CIFAR-100 with a batch size of 64.

\begin{table}[t] \centering
\centering\resizebox{0.25\textwidth}{!}{
\begin{tabular}{c|cc} \toprule
  & \multirow{2}{*}{\shortstack{ResNet32$\times$4 \\ [1pt]ResNet8$\times$4}} & \multirow{2}{*}{\shortstack{VGG13 \\ [1pt]VGG8}} \\ 
  & & \\ \midrule
\cellcolor{sgray} Baseline & \cellcolor{sgray} \textbf{78.76} & \cellcolor{sgray} \textbf{76.19} \\
w/o $L^{S}_{ce}$ & 78.47 & 75.66 \\
\bottomrule
\end{tabular}} 
\caption{Effect of $L^{V}_{ce}$ supervision.} \label{tab:abl_ces}
\end{table}

\subsection{More Experimental Results} \label{sec:appendix_results}
We present the full results on CIFAR-100 in Tabs.~\ref{tab:cifar100_same} and ~\ref{tab:cifar100_diff} to include more same-model and different-model distillation pairs and additional  methods for comparison.

\subsection{More Ablation Studies} \label{sec:appendix_abl}

\paragraph{Effect of GT supervision policies.} 
Tab.~\ref{tab:abl_ces} shows the effect of removing the GT supervision on the student model's predictions of the virtual-view image. It demonstrates that supervising student predictions of the virtual view is important for ensuring the quality of the virtual view predictions. The quality of vertices has a direct impact on the quality of the edges (\textit{i.e.}, relations) constructed within the affinity graphs. As such, we choose to also supervise the virtual vertices of our graphs with GT labels. 

\paragraph{Effect of longer training.}
The construction and transfer of richer and more diverse relations mean that VRM may benefit more from longer training. To verify this, we devise a longer training policy (denoted as ``LT'') than the standard 240-epoch policy in existing KD methods. For our LT policy, the model is trained for 360 epochs and the LR decays by a factor of 10 at the 150th, 180th, 210th, and 270th epochs. All other configurations are kept the same. According to Tab.~\ref{tab:more_epochs}, VRM indeed benefits from longer training as a 0.21\% Top-1 accuracy gain is obtained with LT. In comparison, the performance of other methods plateaued with more training epochs, which is likely due to overfitting to the training samples and a lack of richer guidance signals from the teacher.

\paragraph{Effect of different temperatures.} The temperature of Softmax, denoted by $\tau$, controls the smoothness of the predicted probabilistic distribution. 
We simply opt for the common choice of $\tau=4$~\cite{kd, dkd}, which is empirically shown to produce the best results.  Moreover, the proposed method is sufficiently robust to varying values of $\tau$, as shown in Tab.~\ref{tab:abl_t}. 

\paragraph{Effect of pruning redundant edges.} 
We conduct ablation experiments to see the effect of pruning redundant edges, described in Sec.~\ref{sec:method_pruning} of the main text. As presented in Tab.~\ref{tab:abl_pruning}, matching the raw and bulky inter-sample affinity graph with redundancy and duplication is not only less efficient but also inferior in terms of performance. We postulate that this is partially ascribed the fact that each vertex in the raw graph is connected to a larger and more complex set of other vertices that involve both real and virtual vertices. This complicates the learning while making each vertex more vulnerable to an increased likelihood of adverse gradient propagation. Another possible reason is that matching real-virtual relations is more regularised, as opposed to matching real-real or virtual-virtual intra-view relations that are easier and more readily overfitted. Note that we also conjectured that the degraded performance of matching the raw graph may be ascribed to different distribution patterns of the prediction vectors at the vertices since they now have a dimension of $2\times B$ compared to $B$. We experimented with different temperature $\tau$ in an attempt to re-adjust the distributions to be more relation-matching-friendly, but the results remain inferior.

\begin{table}[t] \centering
\centering\resizebox{0.31\textwidth}{!}{
\begin{tabular}{c|cccc} \toprule
 & KD & RKD & DIST & \cellcolor{sgray} VRM \\ \midrule
Baseline & 73.83 & 72.63 & 76.16 & \cellcolor{sgray} 78.76 \\
LT & 73.82 & 72.49 & 75.90 & \cellcolor{sgray} \textbf{78.97} \\
\bottomrule
\end{tabular}} \caption{Effect of longer training. ``LT'': long-training policy.} \label{tab:more_epochs}
\end{table}

\begin{table}[t] \centering
\centering\resizebox{0.88\linewidth}{!}{
\begin{tabular}{c|cccccc}
\toprule
$\tau$ & 1 & 2 & 3 & \cellcolor{sgray} 4 & 5 & 6 \\ \midrule
Acc. (\%) & 78.16 & 78.69 & 78.57 & \cellcolor{sgray} \textbf{78.76} & 78.41 & 78.63 \\
\bottomrule
\end{tabular}} 
\caption{Effect of different temperatures.}
\label{tab:abl_t}
\end{table}

\subsection{More Analyses} \label{sec:appendix_analyses}
\begin{figure*}[t] \centering
\includegraphics[width=0.80\textwidth]{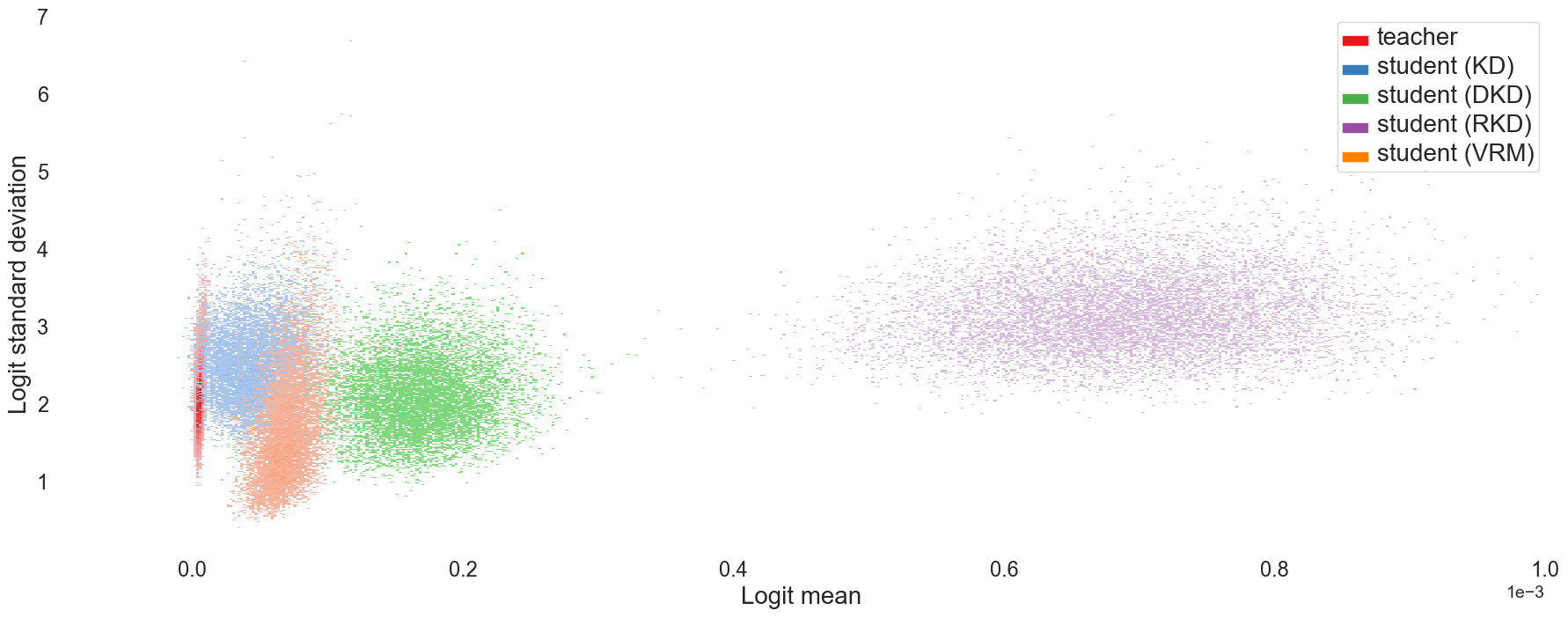}
\caption{Bivariate histogram of the mean and standard deviation of logits predicted by different models on CIFAR-100.} \label{fig:mean_std}
\end{figure*}

\paragraph{Analysis of logit mean \& standard deviation.} In Fig.~\ref{fig:mean_std}, we plot the histogram of the mean and standard deviation of instance-wise logit predictions given by various method. IM methods are found to produce logits closer to the teachers' in terms of both logit mean and standard deviation distributions. Intriguingly, the proposed VRM, being a purely relation-based method that is free of any explicit instance-wise logit matching, is on par with IM methods in this regard and markedly outdoes DKD. This suggests that the proposed real-virtual relational matching provides strong regularisation that better enables the student to learn the underlying logit distribution of the teacher. This is particularly evident given that RKD, a relation-based method which also has an IM objective, is way further from teacher's logit distribution.

\begin{figure}[t!] \centering
\includegraphics[width=0.48\textwidth]{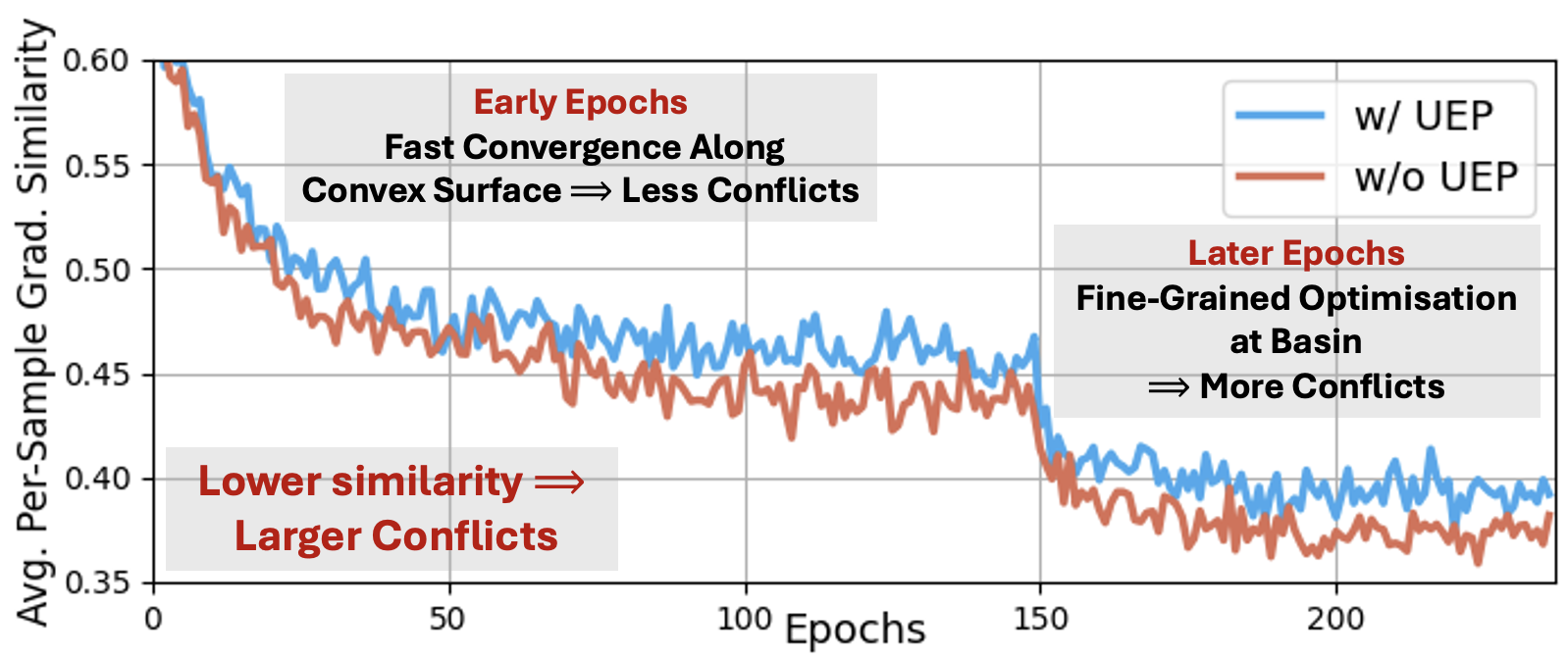}
\caption{Effect of unreliable edge pruning on gradient conflicts in training.} \label{fig:vis_grad_sim}
\end{figure}

\paragraph{Effect of UEP on optimisation conflicts in training.} To gain further insights into the effect of unreliable edge pruning on the training dynamics, we trace the optimisation gradient similarity in matching the edges throughout training. Specifically, we compute the averaged pairwise cosine similarity between all sample-wise gradient vectors within a batch, and visualise the results in Fig.~\ref{fig:vis_grad_sim}. We observe that UEP leads to higher gradient similarity on average. In other words, there are fewer gradient conflicts, which also explains the faster convergence evidenced by Figs~\ref{fig:landscape_vrm}.

\begin{table}[t] \centering
\centering\resizebox{0.33\textwidth}{!}{
\begin{tabular}{c|cc} \toprule
\multirow{2}{*}{\shortstack{Pruning \\ Configuration}}  & \multirow{2}{*}{\shortstack{ResNet32$\times$4 \\ [1pt]ResNet8$\times$4}} & \multirow{2}{*}{\shortstack{VGG13 \\ [1pt]VGG8}} \\ 
& & \\ \midrule
Redundant $\mathcal{E}$ ($\tau =$ 1) & 77.39 & 75.26 \\
Redundant $\mathcal{E}$ ($\tau =$ 2) & 77.80 & 74.94 \\
Redundant $\mathcal{E}$ ($\tau =$ 4) & 77.19 & 75.30 \\
\cellcolor{sgray} Pruned $\mathcal{E}$ ($\tau =$ 4) & \cellcolor{sgray} \textbf{78.76} & \cellcolor{sgray} \textbf{76.19} \\
\bottomrule
\end{tabular}}
\caption{Effect of pruning redundant edges.} \label{tab:abl_pruning}
\end{table}

\paragraph{VRM on features.} In this work, we have chosen to construct our virtual relation graphs $\mathcal{G}^{IS}$ and $\mathcal{G}^{IC}$ from network prediction logits $\{\mathbf{z_i}\}_{i=1}^B$. In this section, we conduct additional experiments to investigate to what extent VRM can work with features. To this end, we simply reconstruct our graphs from the feature maps $\{\mathbf{f}_i\}_{i=1}^B$ right before the final linear layer (denoted as ``pooled feats'' in Tab.~\ref{tab:vrm_feats}) and in the \texttt{mdistiller} codebase. Our virtual relation graphs now become $\mathcal{G}^{IS} \in \mathbb{R}^{B \times B \times D}$ and $\mathcal{G}^{IC} \in \mathbb{R}^{D \times D \times B}$ where D is the dimension of the feature vector. Note that since we no longer work with probability distributions, we remove the Softmax operations that convert predictions to probabilities. Other operations remain unchanged.

In Tab.~\ref{tab:vrm_feats}, we compare the results of VRM trained using graphs constructed from feature maps with existing methods that also build relations from the same features (\textit{i.e.}, ``pooled feats''), namely RKD~\cite{rkd},  PKT~\cite{pkt}, CRD~\cite{crd}, and ReviewKD~\cite{reviewkd}. It can be observed that the performance of VRM deteriorates when applied to features. The reason may be that predicted logits are more compact condensation of categorical knowledge, which is therefore more beneficial for our downstream task. This is particularly so given that VRM does not contain an IM objective that directly matches the logits. As such, VRM works best when applied to logits. Nonetheless, when applied to features, VRM still substantially outperforms all other methods that also work on the very same feature maps. This shows that VRM still encodes better and richer knowledge for distillation compared to the weaker relations transferred by RKD and PKT.

\begin{table}[t] \centering
\centering\resizebox{0.36\textwidth}{!}{
\begin{tabular}{c|c|cc} \toprule
 \multirow{2}{*}{Method} &  \multirow{2}{*}{Location} & \multirow{2}{*}{\shortstack{ResNet32$\times$4 \\ [1pt]ResNet8$\times$4}} & \multirow{2}{*}{\shortstack{VGG13 \\ [1pt]VGG8}} \\ 
 & & & \\ \midrule
RKD & \multirow{5}{*}{pooled feats} & 72.63 & 70.87 \\
PKT &  & 74.41 & 72.78 \\
CRD &  & 75.51 & 73.94 \\
ReviewKD &  & 75.63 & 74.84 \\
\cellcolor{sgray} VRM &  & \cellcolor{sgray} \textbf{76.39} & \cellcolor{sgray} \textbf{74.92} \\ \midrule
VRM & logits & \textbf{78.76} & \textbf{76.19} \\
\bottomrule
\end{tabular}} \caption{Applying VRM to feature embeddings.} \label{tab:vrm_feats}  
\end{table}

\paragraph{Discussions on the use of joint entropy.}
In the formulation of our unreliable edge pruning scheme, we use the joint entropy (JE) between two predictions (of two nodes) as a measure of edge uncertainty. 
While other measures may be used, JE suits our purpose with several appealing properties: \vspace{1mm}
\begin{itemize}
    \item Higher discrepancy between two vertices leads to higher JE, which is a relative measure of uncertainty. \vspace{1mm}
    \item As two predictions get aligned, JE approaches their individual uncertainty, which is an absolute measure of uncertainty.
\end{itemize} \vspace{1mm}
As such, our criterion takes account of both relative and absolute edge uncertainties throughout the learning process.

\paragraph{Comparison to KD methods using SSL.} We highlight the difference between our method and two KD methods that utilise self-supervised learning (SSL), namely SSKD~\cite{sskd} and HSAKD~\cite{hsakd}.

SSKD utilises self-supervision signals via image transformations and pretext tasks for knowledge distillation. The proposed VRM fundamentally differs from SSKD in at least the following aspects: 

\begin{itemize}
\item \textit{Motivation}: While both methods involve the utilisation of transformations to produce augmented views of input images, SSKD is directly inspired by and leverage the \textit{pretext task} in self-supervised learning~\cite{simclr}. In contrast, our method is pretext-task-free. Instead, VRM is motivated by \textit{transformation invariance regularisation} which was originally popularised in semi-supervised learning~\cite{pea, piemodel} and domain adaptation~\cite{adamatch}. \vspace{1mm}

\item \textit{Training formulation}: A direct consequence of the point above is that SSKD's teacher first needs to be re-trained with additional augmentations (which also causes SSKD to use teachers of higher accuracy than ours), followed by a separate fine-tuning stage for the pretext task. These lead to significantly more procedures and computations, whereas VRM is entirely free of such palaver. \vspace{1mm}

\item \textit{Nature of matching objectives}: SSKD is essentially a \textit{hybrid} method that employs both relation matching and instance-to-instance matching objectives, whereas VRM is \textit{purely relation-based} method. In other words, SSKD relies on IM to achieve competitive performance, while VRM involves purely relation-based objectives. \vspace{1mm}

\item \textit{Design choices}: The designs of both methods are vastly different, including but not limited to the formulation of relations and the choices of augmentation policies, relation distance metrics, and model outputs used for computing relations.
\end{itemize}

\begin{figure*}[t]
    \centering 
    \begin{minipage}{0.33\textwidth}
        \centering
        \includegraphics[width=\textwidth, height=0.88\textwidth]{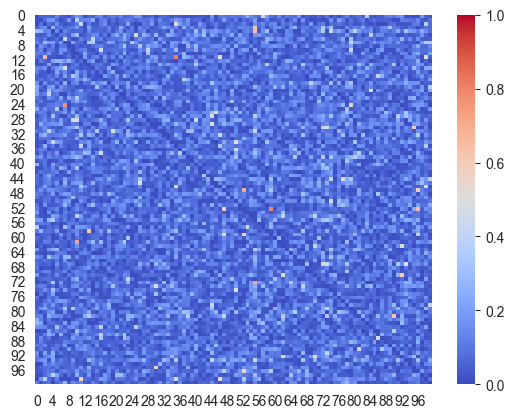}
        \subcaption{KD}
        \label{fig:subfig1}
    \end{minipage}\hfill
    \begin{minipage}{0.33\textwidth}
        \centering
        \includegraphics[width=\textwidth, height=0.88\textwidth]{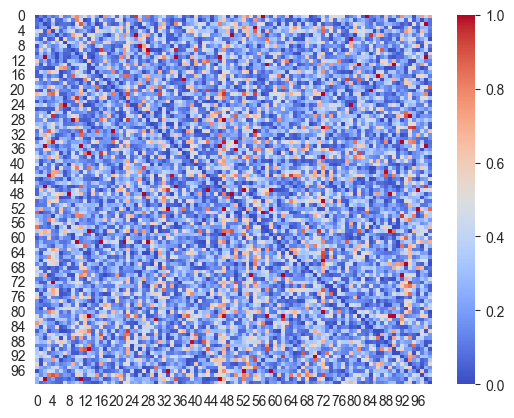}
        \subcaption{RKD}
        \label{fig:subfig2}
    \end{minipage} \hfill
    \begin{minipage}{0.33\textwidth}
        \centering
        \includegraphics[width=\textwidth, height=0.88\textwidth]{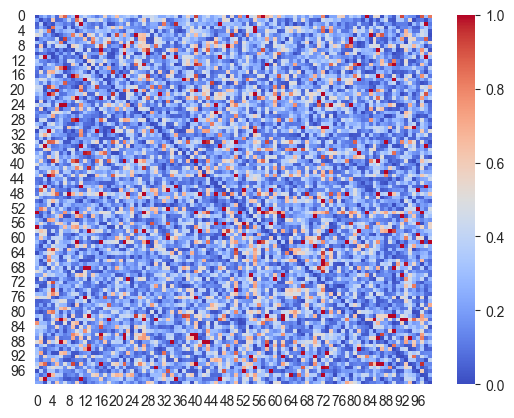}
        \subcaption{PKT}
        \label{fig:subfig2}
    \end{minipage}
    \begin{minipage}{0.33\textwidth}
        \centering
        \includegraphics[width=\textwidth, height=0.88\textwidth]{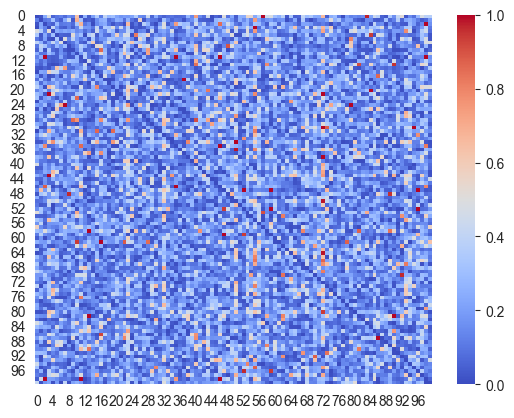}
        \subcaption{SP}
        \label{fig:subfig3}
    \end{minipage}\hfill
    \begin{minipage}{0.33\textwidth}
        \centering
        \includegraphics[width=\textwidth, height=0.88\textwidth]{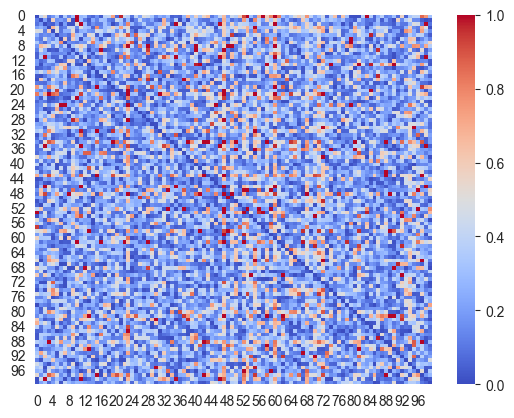}
        \subcaption{ICKD}
        \label{fig:subfig4}
    \end{minipage} \hfill
    \begin{minipage}{0.33\textwidth}
        \centering
        \includegraphics[width=\textwidth, height=0.88\textwidth]{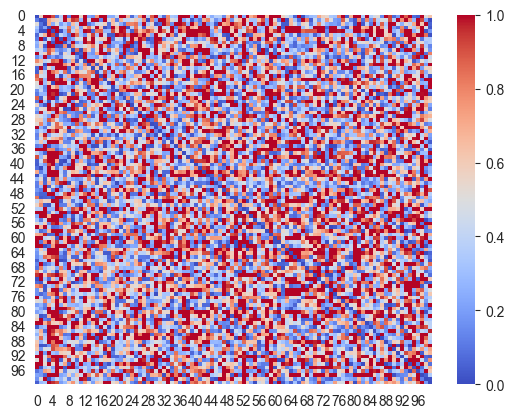}
        \subcaption{DIST}
        \label{fig:subfig4}
    \end{minipage}
    \begin{minipage}{0.33\textwidth}
        \centering
        \includegraphics[width=\textwidth, height=0.88\textwidth]{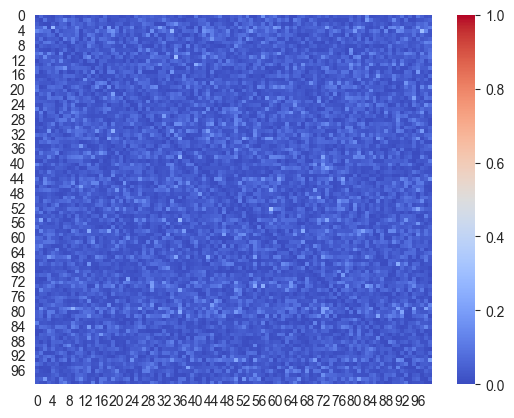}
        \subcaption{DKD}
        \label{fig:subfig3}
    \end{minipage}\hfill
    \begin{minipage}{0.33\textwidth}
        \centering
        \includegraphics[width=\textwidth, height=0.88\textwidth]{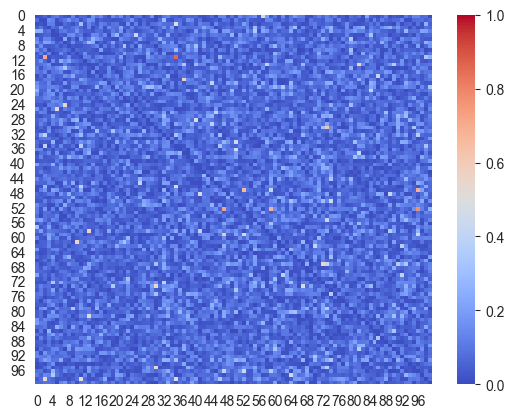}
        \subcaption{MLLD}
        \label{fig:subfig4}
    \end{minipage}
    \begin{minipage}{0.33\textwidth}
        \centering
        \includegraphics[width=\textwidth, height=0.88\textwidth]{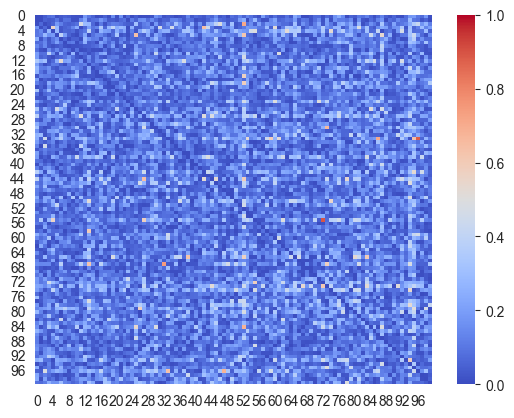}
        \subcaption{VRM}
        \label{fig:subfig4}
    \end{minipage}
    \caption{More visualisations of teacher-student prediction discrepancy maps for different KD methods.} \label{fig:appendix_discrepancy}
\end{figure*} 

\vspace{1mm}
HSAKD is another method that makes use of self-supervised learning and transformed views of input images. This method is also fundamentally different from VRM from the following aspects: 

\begin{itemize}
\item \textit{Motivation}:
Like SSKD, HSAKD is also directly motivated by the use of pretext tasks in self-supervised learning. HSAKD employs rotation prediction as its pretext task. The proposed VRM is free of pretext task learning. \vspace{1mm}

\item \textit{Training formulation}: To enable pretext task learning, HSAKD appends auxiliary classifiers to the intermediate features at each stage to perform transformation classification. This means that, akin to SSKD, HSAKD also needs to re-train the teacher model with modified architecture over the pretext task. The auxiliary classifiers also introduce extra parameters. In contrast, the proposed VRM does not involve these additional procedural, parameter, and computational costs. \vspace{1mm}

\item \textit{Nature of matching objectives}: By matching the predictions made by a set of auxiliary classifiers between teacher and student for each sample (as well as matching the final predicted probability distributions between teacher and student), HSAKD is fundamentally a instance matching approach, whereas VRM transfers purely relational knowledge.  Moreover, HSAKD employs symmetric matching, which means the matching between teacher and student auxiliary predictions are for the same view of the input samples. By contrast, VRM exploits the relations across asymmetric real and virtual views with different difficulties. \vspace{1mm}

\item \textit{Design choices}: HSAKD also differs from the proposed VRM in terms of specific designs made. For example, HSAKD adopts rotation to construct its pretext task, whereas VRM utilises RandAugment policies. HSAKD also relies on the use of the instance-wise logit matching loss from vanilla KD~\cite{kd} to reach competitive performance, whereas VRM does not use any instance-matching KD objective and still achieves much more superior performance.
\end{itemize}

\begin{figure*}[t] \vspace{-4mm}
    \centering 
    \begin{minipage}{0.33\textwidth}
        \centering
        \includegraphics[width=\textwidth, height=0.9\textwidth]{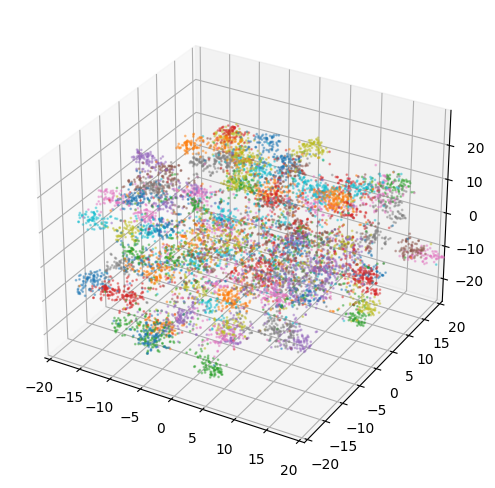}
        \subcaption{KD}
        \label{fig:subfig1}
    \end{minipage}\hfill
    \begin{minipage}{0.33\textwidth}
        \centering
        \includegraphics[width=\textwidth, height=0.9\textwidth]{figures/tsne_rkd_3d.png}
        \subcaption{RKD}
        \label{fig:subfig2}
    \end{minipage} \hfill
    \begin{minipage}{0.33\textwidth}
        \centering
        \includegraphics[width=\textwidth, height=0.9\textwidth]{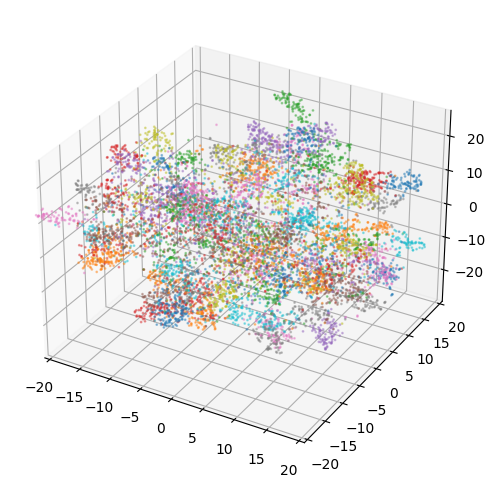}
        \subcaption{PKT}
        \label{fig:subfig2} 
    \end{minipage}
    \begin{minipage}{0.33\textwidth}
        \centering
        \includegraphics[width=\textwidth, height=0.9\textwidth]{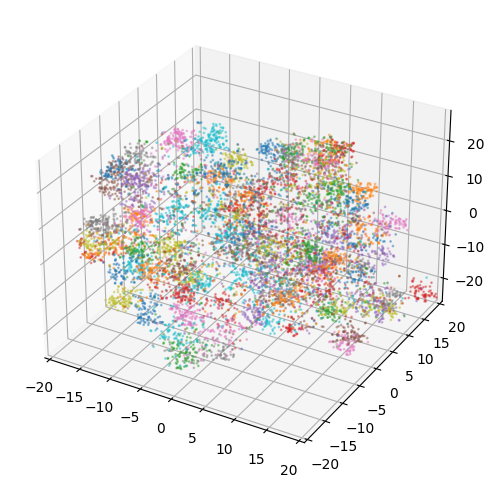}
        \subcaption{SP}
        \label{fig:subfig3}
    \end{minipage}\hfill
    \begin{minipage}{0.33\textwidth}
        \centering
        \includegraphics[width=\textwidth, height=0.9\textwidth]{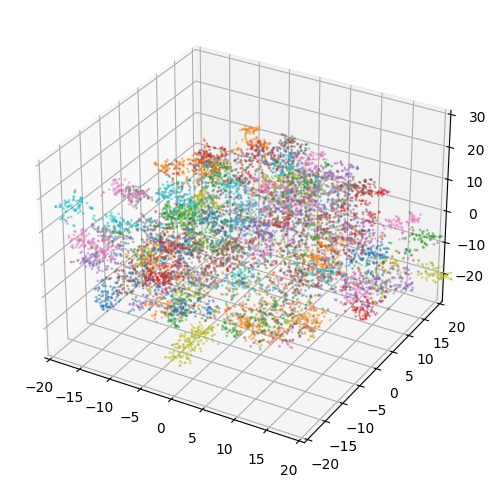}
        \subcaption{ICKD}
        \label{fig:subfig4}
    \end{minipage} \hfill
    \begin{minipage}{0.33\textwidth}
        \centering
        \includegraphics[width=\textwidth, height=0.9\textwidth]{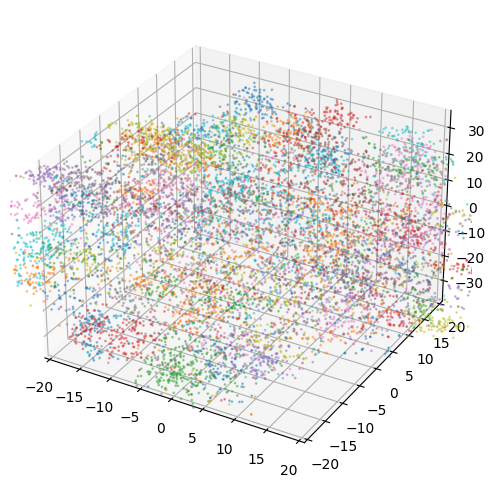}
        \subcaption{DIST}
        \label{fig:subfig4}
    \end{minipage}
    \begin{minipage}{0.33\textwidth}
        \centering
        \includegraphics[width=\textwidth, height=0.9\textwidth]{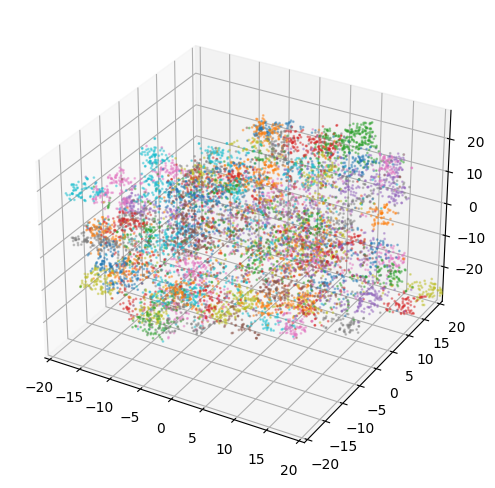}
        \subcaption{FitNets}
        \label{fig:subfig3}
    \end{minipage}\hfill
    \begin{minipage}{0.33\textwidth}
        \centering
        \includegraphics[width=\textwidth, height=0.9\textwidth]{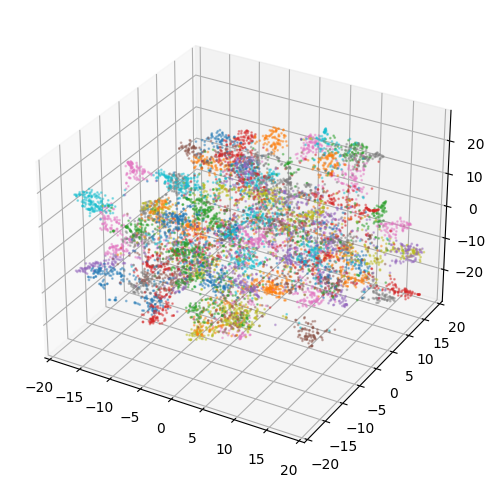}
        \subcaption{DKD}
        \label{fig:subfig4}
    \end{minipage}
    \begin{minipage}{0.33\textwidth}
        \centering
        \includegraphics[width=\textwidth, height=0.9\textwidth]{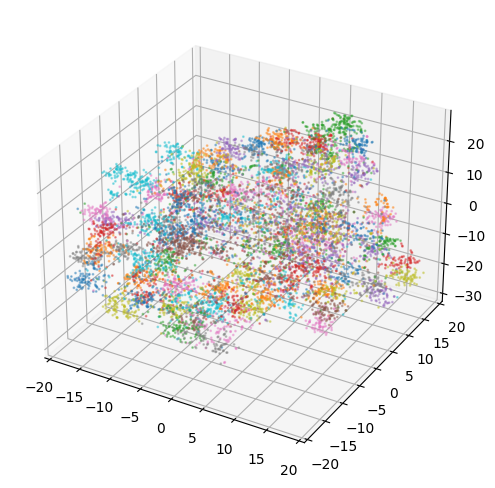}
        \subcaption{MLLD}
        \label{fig:subfig4}
    \end{minipage}
    \begin{minipage}{0.33\textwidth}
        \centering
        \includegraphics[width=\textwidth, height=0.9\textwidth]{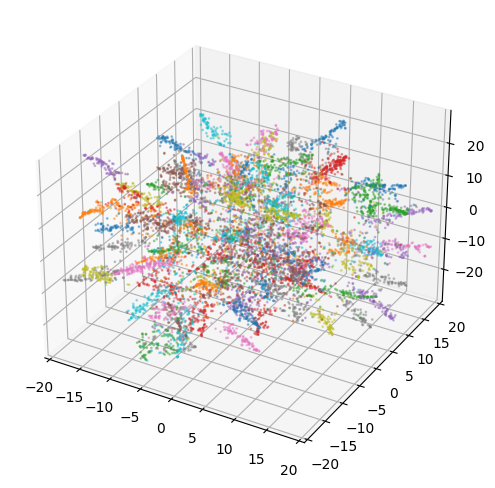} 
        \subcaption{LSKD}
        \label{fig:subfig3}
    \end{minipage}\hfill
    \begin{minipage}{0.33\textwidth}
        \centering
        \includegraphics[width=\textwidth, height=0.9\textwidth]{figures/tsne_vrm_3d.png}
        \subcaption{VRM}
        \label{fig:subfig4}
    \end{minipage}
    \begin{minipage}{0.33\textwidth}
        \centering
        \includegraphics[width=\textwidth, height=0.9\textwidth]{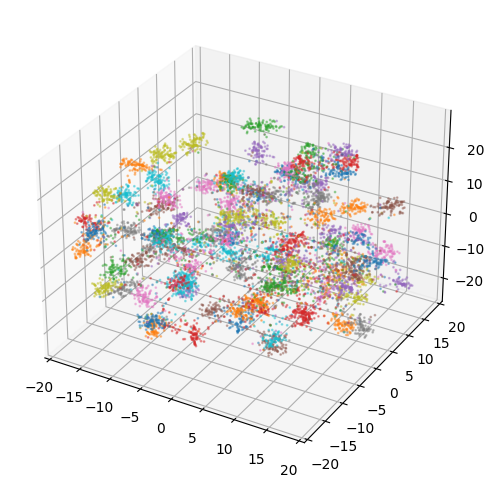}
        \subcaption{Teacher}
        \label{fig:subfig4}
    \end{minipage}
    \vspace{-1mm}
    \caption{More t-SNE visualisations of features learnt by different KD methods.} \label{fig:appendix_tsne} \vspace{-6mm}
\end{figure*}

\begin{figure*}[t] \vspace{-60mm}
    \centering
    \begin{minipage}[t]{0.32\textwidth}
        \centering
        \includegraphics[width=\textwidth]{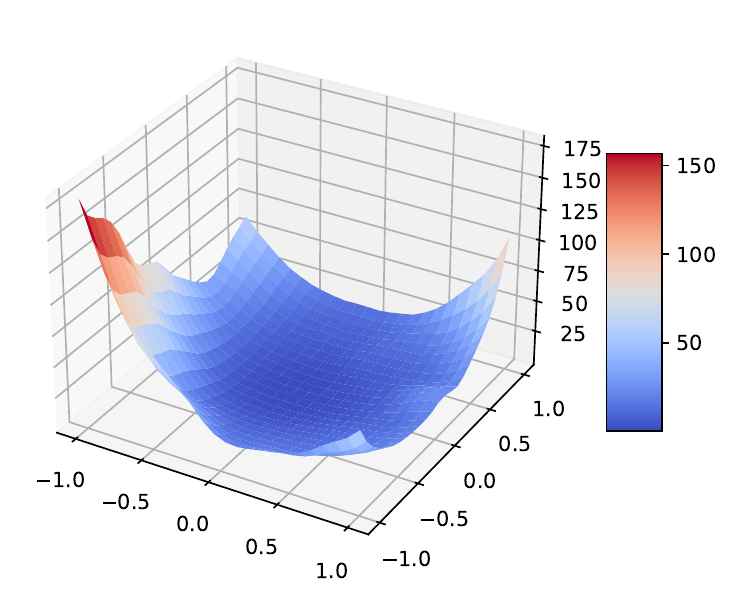}
        \subcaption{KD}
        \label{fig:subfig1}
    \end{minipage}\hfill
    \begin{minipage}[t]{0.32\textwidth}
        \centering
        \includegraphics[width=\textwidth]{figures/rkd_r8x4.pdf}
        \subcaption{RKD}
        \label{fig:subfig2}
    \end{minipage} 
    \begin{minipage}[t]{0.32\textwidth}
        \centering
        \includegraphics[width=\textwidth]{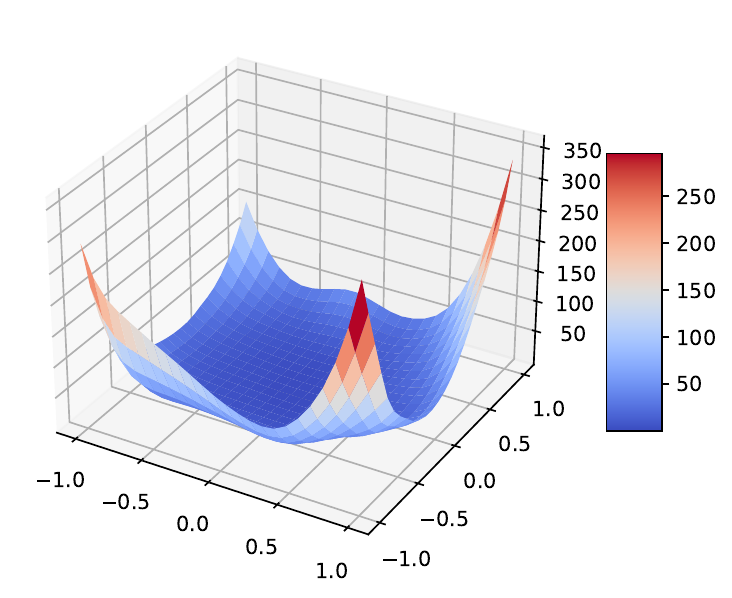}
        \subcaption{PKT}
        \label{fig:subfig3}
    \end{minipage}
    \vspace{0.5em}
    \begin{minipage}[t]{0.32\textwidth}
        \centering
        \includegraphics[width=\textwidth]{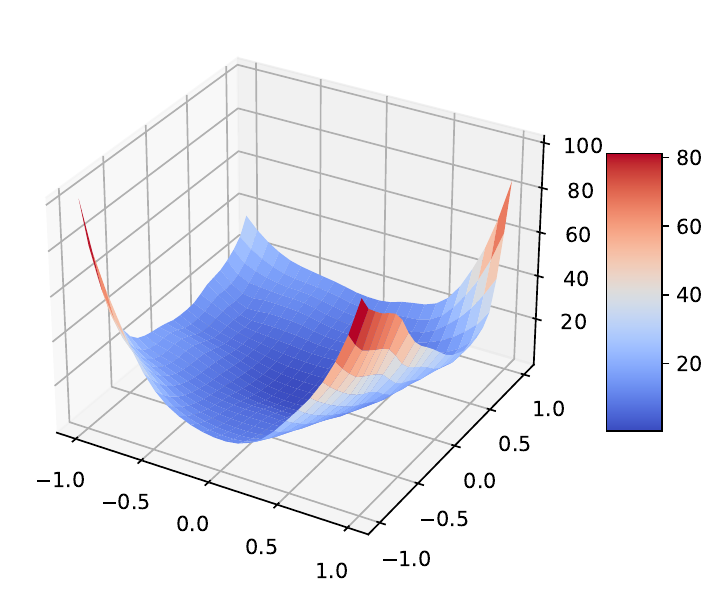}
        \subcaption{SP}
        \label{fig:subfig4}
    \end{minipage} \hfill
    \begin{minipage}[t]{0.32\textwidth}
        \centering
        \includegraphics[width=\textwidth]{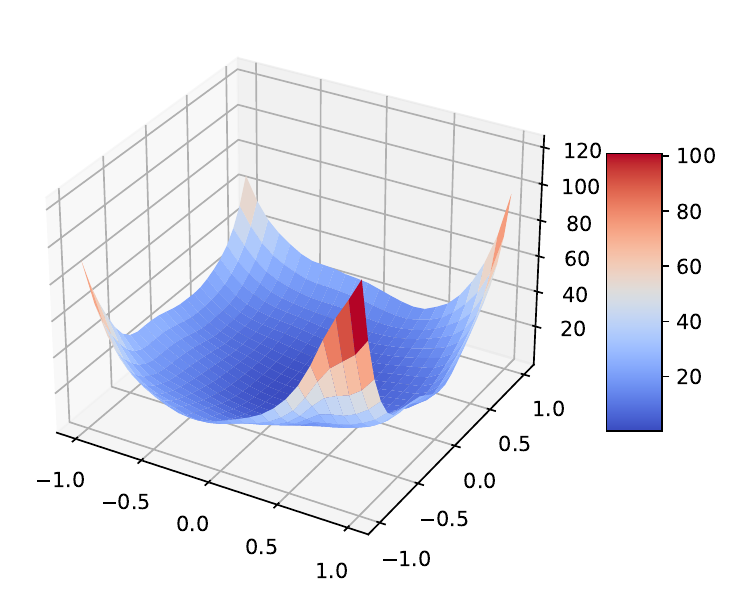}
        \subcaption{ICKD}
        \label{fig:subfig3}
    \end{minipage}\hfill
    \begin{minipage}[t]{0.32\textwidth}
        \centering
        \includegraphics[width=\textwidth]{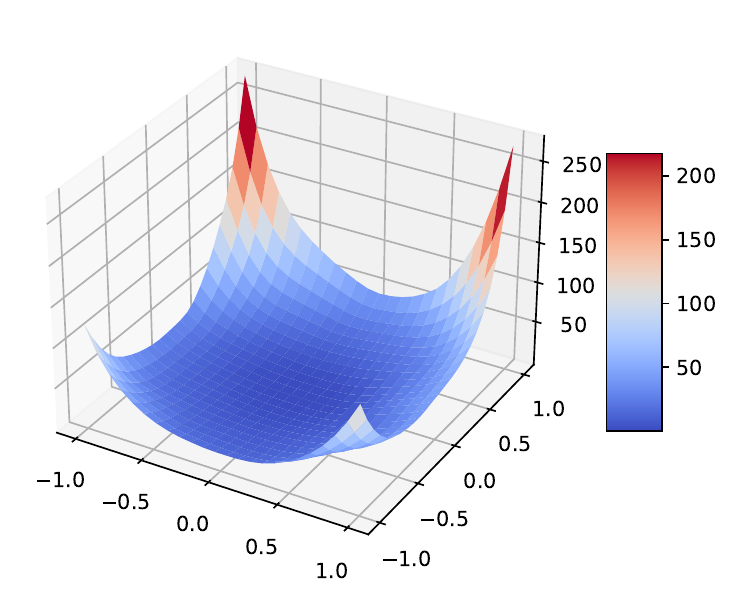}
        \subcaption{DIST}
        \label{fig:subfig4}
    \end{minipage}
    \vspace{0.5em}
    \begin{minipage}[t]{0.32\textwidth}
        \centering
        \includegraphics[width=\textwidth]{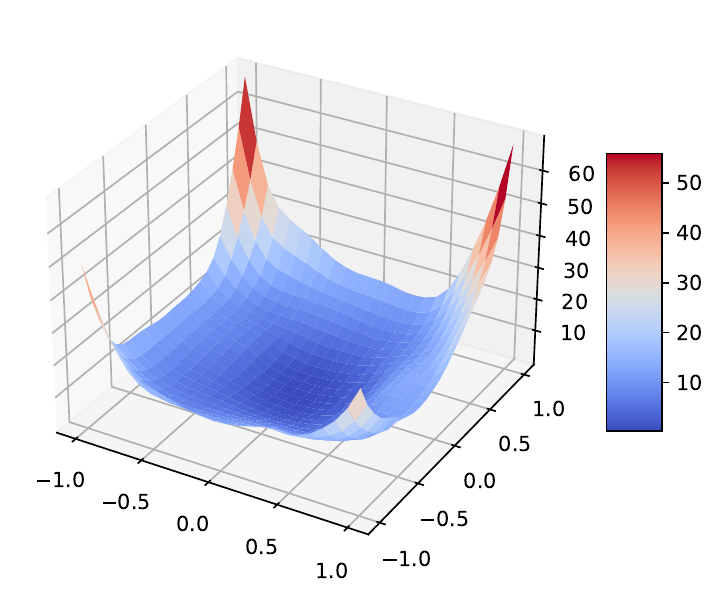}
        \subcaption{DKD}
        \label{fig:subfig4}
    \end{minipage} \hfill
    \begin{minipage}[t]{0.32\textwidth}
        \centering
        \includegraphics[width=\textwidth]{figures/vrm_r8x4.pdf}
        \subcaption{VRM}
        \label{fig:subfig4}
    \end{minipage} \hfill
    \begin{minipage}[t]{0.32\textwidth}
        \centering
        \includegraphics[width=\textwidth]{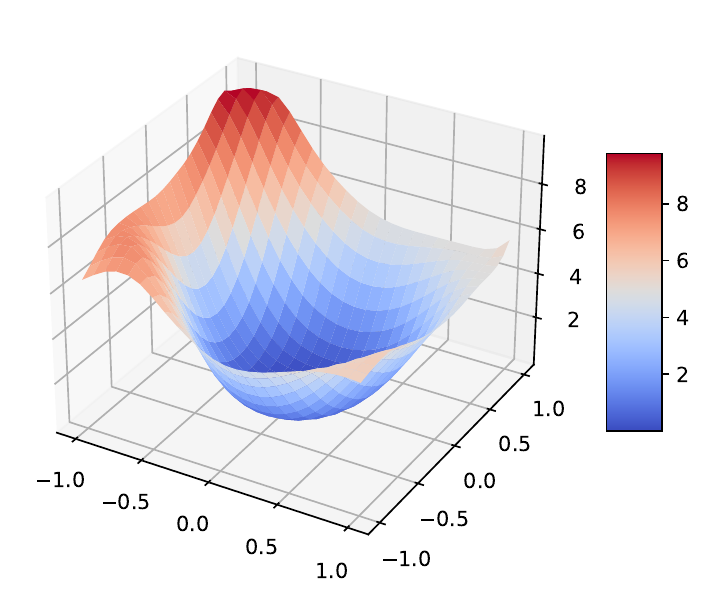}
        \subcaption{Teacher}
        \label{fig:subfig4}
    \end{minipage}
    \caption{More loss landscape visualisations of different KD methods.}
    \label{fig:appendix_landscape}
\end{figure*}

\paragraph{More teacher-student  discrepancy maps visualisations.} Fig.~\ref{fig:appendix_discrepancy}, we provide more visualisations of the class-wise prediction discrepancy between teacher and student models for different KD methods (ResNet32$\times$4 $\!\to\!$ ResNet8$\times 4$ on CIFAR-100).

\paragraph{More t-SNE visualisations.} In Fig.~\ref{fig:appendix_tsne}, we further showcase the t-SNE visualisations of embeddings learnt by more KD methods as well as those by the teacher model (ResNet32$\times$4 $\!\to\!$ ResNet8$\times 4$ on CIFAR-100).

\paragraph{More loss landscape visualisations.}
Fig.~\ref{fig:appendix_landscape} provides more visualisations of loss landscape for different KD methods (ResNet32$\times$4 $\!\to\!$ ResNet8$\times 4$ on CIFAR-100).

\end{document}